\documentclass[10pt,twocolumn,letterpaper]{article}

\usepackage{cvpr}
\usepackage{times}
\usepackage{epsfig}
\usepackage{graphicx}
\usepackage{amsmath}
\usepackage{amssymb}
\usepackage{overpic}
\usepackage{xcolor}
\usepackage{tabu}

\usepackage{pict2e} 
\renewcommand{\vec}[1]{\boldsymbol{#1}}
\newcommand{\mat}[1]{\mathbf{#1}}

\newcommand{\blendweights}[0]{\mat{W}}

\newcommand{\pose}[0]{\vec{\theta}}
\newcommand{\shape}[0]{\vec{\beta}}

\newcommand{\trans}[0]{\vec{t}}

\newcommand{\offsets}[0]{\mathbf{D}}

\newcommand{\posefun}[0]{T}
\newcommand{\blendfun}[0]{W}

\newcommand{\jointfun}[0]{J}

\usepackage[pagebackref=true,breaklinks=true,letterpaper=true,colorlinks,bookmarks=false]{hyperref}

\cvprfinalcopy
\def\cvprPaperID{6177}

\begin{document}
\title{Learning to Transfer Texture from Clothing Images to 3D Humans}
\author{Aymen Mir\textsuperscript{1} \qquad Thiemo Alldieck\textsuperscript{1, 2} \qquad Gerard Pons-Moll\textsuperscript{1}\\\\
{\small \textsuperscript{1}Max Planck Institute for Informatics, Saarland Informatics Campus, Germany}\\
{\small\textsuperscript{2}Computer Graphics Lab, TU Braunschweig, Germany}\\
{\tt\scriptsize \{amir,gpons\}@mpi-inf.mpg.de alldieck@cg.cs.tu-bs.de}}

\makeatletter
\let\@oldmaketitle\@maketitle
\renewcommand{\@maketitle}{
	\@oldmaketitle
	\centering
	\vspace{-4mm}
	\includegraphics[width=\textwidth]{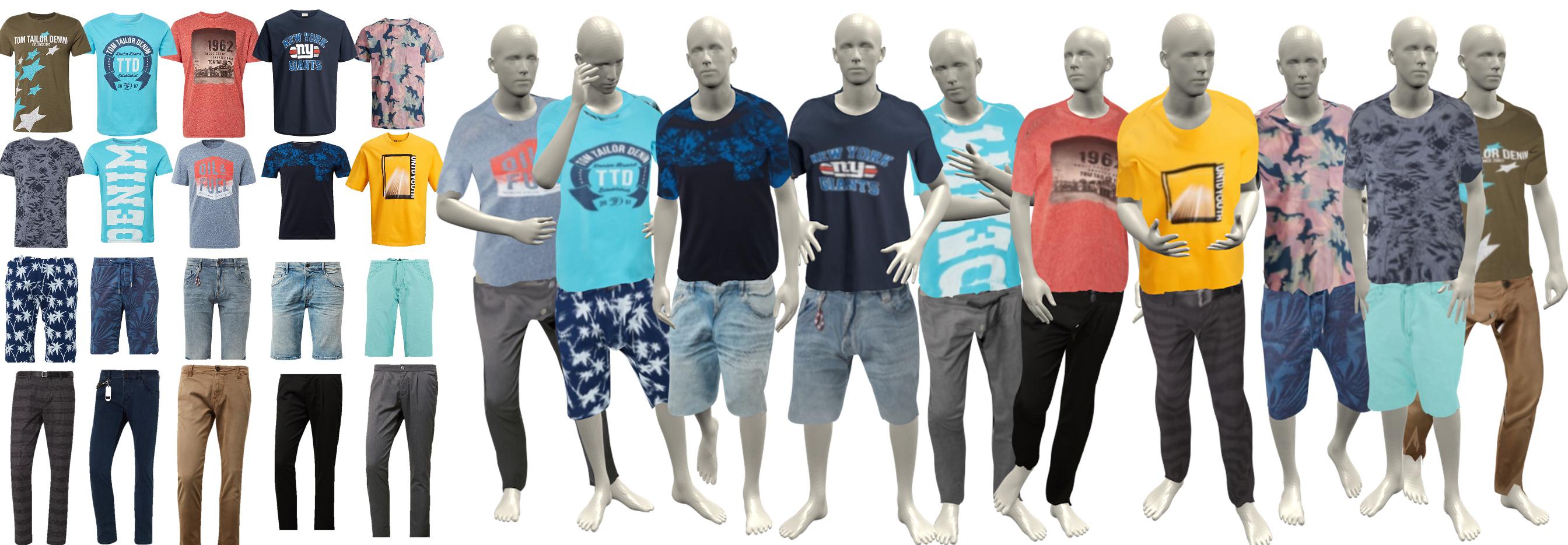}
	\\
	\vspace{2mm}
	\refstepcounter{figure}\normalfont Figure~\thefigure: Our model Pix2Surf allows to digitally map the texture of online retail store clothing images to the 3D surface of virtual garment items enabling 3D virtual try-on in real-time.
	\label{fig:teaser}
	\vspace{5mm}
}
\makeatother

\maketitle

\begin{abstract}
\vspace{-0.5cm}
In this paper, we present a simple yet effective method to automatically transfer textures of clothing images (front and back) to 3D garments worn on top SMPL~\cite{smpl2015loper}, in real time. 
We first automatically compute training pairs of images with aligned 3D garments using a custom non-rigid 3D to 2D registration method, which is accurate but slow. 
Using these pairs, we learn a mapping from pixels to the 3D garment surface. 
Our idea is to learn dense correspondences from garment image silhouettes to a 2D-UV map of a 3D garment surface using shape information alone, completely ignoring texture, which allows us to generalize to the wide range of web images.
Several experiments demonstrate that our model is more accurate than widely used baselines such as thin-plate-spline warping and image-to-image translation networks while being orders of magnitude faster. Our model opens the door for applications such as virtual try-on, and allows for generation of 3D humans with varied textures which is necessary for learning. Code will be available at https://virtualhumans.mpi-inf.mpg.de/pix2surf/.
\end{abstract}
\section{Introduction}
\label{sec:introduction}
Our goal is to learn a model capable of transferring texture from two photographs (front and back) of a garment to the 3D geometry of a garment template, automatically and in real time. 
Such a model can be extremely useful for photo-realistic rendering of humans, varied generation of synthetic data for learning, virtual try-on, art and design.
Despite the many applications, automatic transfer of clothing texture to 3D garments has received very little attention.
The vast majority of recent methods work in image space directly. 
Most works focus on either image based person re-posing~\cite{zhao2018multi,ma2017pose,ma2018disentangled,Pumarola_cvpr2018a,siarohin2018deformable,dong2018soft,balakrishnan2018synthesizing,song2019unsupervised}, or virtual try-on \cite{raj2018swapnet,zanfir2018human,han2017viton,wang2018toward,Yu_2019_ICCV,Dong_2019_Towards,zheng2019virtually}. Re-posing methods learn to synthesize image pixels to produce novel poses of a person, whereas virtual try-on methods learn to morph an image of a clothing item to drape it on a target person. The advantage of these methods is that they can be trained on large-scale datasets. The critical disadvantage is that they operate in pixel-space instead of 3D, they can not synthesize difficult poses, and struggle to produce temporally consistent results.
Another line of work extracts texture by fitting 3D human models (SMPL~\cite{smpl2015loper}) to images~\cite{bhatnagar2019mgn,alldieck2018video,alldieck2019learning}, but texture quality quickly deteriorates for complex poses. Other works map texture to 3D meshes~\cite{varol17_surreal,moulding,lazova3dv2019,ponsmoll2017clothcap} by 3D scanning people, but the number and variety of clothing textures is limited because 3D scanning is time consuming. 
To break the lack of 3D data barrier, our idea is to learn a dense mapping from images of clothing items, which are ubiquitous on the internet, to the 3D surface of a parametric garment template directly~\cite{bhatnagar2019mgn}.
However, accurate texture transfer is far from trivial: web images vary in texture, garment size and style, pose, and background. 
Nonetheless, in comparison to clothing worn by humans, web-store images have less variation, which we exploit for learning our model. 
Instead of manually annotating garment landmarks~\cite{liu2016fashionlandmark,alp2018densepose}, our idea is to collect training pairs by non-rigidly aligning a parametric 3D garment template to images. 
We leverage the parameterized garment models of MGN~\cite{bhatnagar2019mgn}, non-rigidly fit their surface to image silhouettes. 
While our alignment usually produces good results, it is slow ($~5-15$ minutes per image), and fails in $~5\%$ of the cases. 
Consequently, using only the successful fits, we learn a direct mapping from image pixels which runs in milliseconds, and is more robust than the optimization based approach. 
Our key hypothesis is that the \emph{mapping is determined by the image silhouette shape alone}, and not by appearance. 
Hence, we train a CNN to predict correspondences from a UV map of the garment surface to pixel locations, given the silhouette shape alone as input. Since the model learns in-variances to shape and pose as opposed to appearance, it generalizes to a wide variety of garment images of varied textures. 
We refer to the correspondence predictor as Pix2Surf.
\begin{figure}
	\centering
    \includegraphics[width=\linewidth]{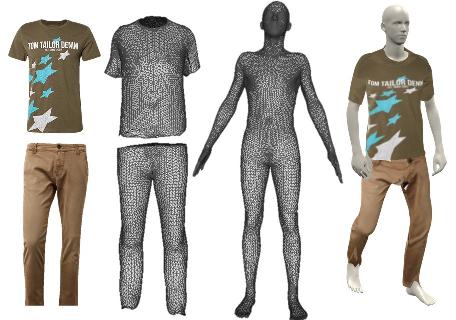}
    \caption{Given regular online retail store images, our method can automatically produce textures for pre-defined garment templates, that can be used to virtually dress SMPL~\cite{smpl2015loper} in 3D.}
    \label{fig:dressing_smpl}
\end{figure}
Pix2Surf allows to digitally map the texture of a clothing image to the 3D surface of a garment in real time, see Fig.~\ref{fig:teaser}.
Once the texture is mapped to 3D, we generalize to novel views, shapes and pose by applying 3D transformations to the 3D geometry~\cite{bhatnagar2019mgn} as illustrated in Fig.~\ref{fig:dressing_smpl}.
Pix2Surf enables, for the first time, virtual try-on from images in 3D directly and in real time, enabling applications such as VR/AR, gaming, and 3D content production. 

To stimulate further research in this direction, we will make Pix2Surf publicly available for research purposes. 
This will allow researchers and users to add texture to the SMPL+Garments model~\cite{bhatnagar2019mgn}, generate synthetic humans, and visualize garment images in 3D directly.  

\section{Related Work}
\label{sec:related}
Synthesizing images of people in realistic clothing is a long-studied problem in the Computer Graphics community. Early works enable to represent and parametrize fabric virtually in 3D \cite{weil1986synthesis,ng1996computer}, later realistic draping and animation has been achieved \cite{baraff1998large,house2000cloth}. While these works require artist-designed garments, careful parameter selection, and computational expensive physics-based simulation, fully-automated and faster methods have been introduced more recently. In contrast, these methods process sensor data such as images, depth maps, or 3D point clouds, and produce 3D reconstructions or photorealistic images with minimal interaction or even fully automatically.\newline
\textbf{2D image synthesis} methods produce images of people holding a given pose or wearing a desired outfit.
These works utilize recent advances in conditional image-to-image translation using generative adversarial networks (GANs) \cite{goodfellow2014generative,isola2017pix2pix}. E.g. \cite{Lassner:GP:2017} presents a method to produce full-body images of people in clothing from semantic segmentation of the body and clothing.
However, the produced subject and outfit are not controllable. The method in \cite{zhao2018multi} produces novel views of given fashion images. While the subject and outfit remain untouched, the pose can be constrained with a sparse set of camera angles. To gain more control over the output, a large number of works deal with synthesizing person images under a desired pose \cite{ma2017pose,ma2018disentangled,Pumarola_cvpr2018a,siarohin2018deformable,dong2018soft,balakrishnan2018synthesizing,song2019unsupervised,shysheya2019textured}. For the same purpose, in \cite{grigorev2019coordinate}, the authors learn warpings between images and the SMPL texture. Predicting correspondences instead of color values helps the network to generalize due to much lower variation in the warps than in the color-space.
In Pix2Surf we also infer correspondences \cite{shotton2012efficient,Pons-Moll_MRFIJCV,alp2018densepose,neverova2018dense} but focus on garments rather than on full bodies. In similar works, \cite{neverova2018dense} utilizes DensePose \cite{alp2018densepose} to warp image pixels into the SMPL texture, but inpainting is performed before warping into the target pose.
More related to our work, recent methods focus on exchanging the a subject's outfit while preserving \cite{raj2018swapnet,zanfir2018human} or changing \cite{han2017viton,wang2018toward,Yu_2019_ICCV,Dong_2019_Towards,zheng2019virtually} his or her pose.
Additionally, special case methods for text-based fashion image manipulation \cite{zhu2017your} and image-based outfit improvement \cite{Hsiao_2019_ICCV} have been presented.
In contrast to our method, all these \emph{virtual try-on} methods work in the image-space and thus perform no explicit reasoning about the underlying 3D scene. This means they are not guaranteed to produce consistent output under varying poses and camera angles. This is approached in \cite{Dong_2019_ICCV} via image warping and subsequent refinement of a previously generated image. While 2D warping improves the quality of synthesized videos, limitations of 2D methods are still present. \newline
\textbf{3D reconstruction} methods focus on recovering the actual 3D shape of a captured garment alone, the body shape of a subject wearing the garment, or both simultaneously.
Methods utilizing controlled RGB \cite{jeong2015garment} and RGB-D images \cite{chen2015garment} have been presented, that select and refine 3D garment templates based on image observations.
While \cite{jeong2015garment} utilizes a tailor's dummy, \cite{chen2015garment} expects the subject to hold a certain pose. Other methods focus on recovering the shape and detailed garment wrinkles of clothing item in less controlled settings \cite{popa2009wrinkling,danvevrek2017deepgarment,bednarik2018learning,jin2018pixel,lahner2018deepwrinkles}.
While these methods can produce detailed geometry, none of these methods focuses on the appearance of the item.
Another branch of research aims at 3D reconstructing the whole human including clothing.
This can be achieved by optimization-based \cite{alldieck2018video,alldieck2018detailed,weng2018photo} or learning-based methods \cite{alldieck2019learning,lazova3dv2019} that utilize silhouettes or semantic segmentation of a short monocular video clip or recently even from single images \cite{natsume2018siclope,alldieck2019tex2shape,pifuSHNMKL19,moulding} and point-clouds \cite{chibane20ifnet}.
Other methods utilize Kinect-fusion like approaches \cite{izadi2011kinectfusion,newcombe2011kinectfusion} to scan people using RGB-D sensors \cite{shapiro2014rapid,3Dportraits,zeng2013templateless,cui2012kinectavatar}.
Having a 3D reconstruction of body and clothing, it can be used to non-rigidly track the subject \cite{MonoPerfCap_SIGGRAPH2018,Habermann:2019:LiveCap}.
All these methods fuse body and clothing in a single template.
Virtual try-on applications, however, often require separate meshes \cite{hauswiesner2011free}.
Therefore, methods that reconstruct the naked body shape or both body shape and clothing have been developed.
The naked body shape alone has been estimated using several RGB images  \cite{bualan2008naked} or more accurately using a sequence of clothed scans \cite{zahng2017shapeundercloth}.
Body shape and garments have been reconstructed simultaneously and fully-automatically from a series of scans \cite{ponsmoll2017clothcap}, RGB-D images \cite{SimulCap19}, and recently even from a small set of RGB images \cite{bhatnagar2019mgn}.
In \cite{yang2018physics} the authors present garment and body-shape recovery from a single image but heavily rely on physical priors and human interaction.
In order to enable dynamic virtual try-on and clothing re-targeting, joint models of clothing and the human body have been developed \cite{neophytou2014layered,yang2018analyzing,patel20vtailor,ma20autoenclother}.
Again, all these works focus mainly or exclusively on the geometry of the garment, not on their appearance. 
Other works also learn to predict correspondences from depth maps to surfaces~\cite{VM,Pons-Moll_MRFIJCV,wei2016dense}, image to surfaces~\cite{kulkarni2019canonical,alp2018densepose,grigorev2019coordinate}, 
but they all address different problems. \newline
\textbf{Automatic texturing} of 3D models from photos has been presented too, but the shape has to be first aligned with the input image \cite{Park:2018:PPM:3272127.3275066,Wang:2016:UTT:2980179.2982404}.
This alignment is expensive and error-prone as silhouette and feature cues may be ambiguous.
The most related work here is ~\cite{stanford_3DV}, which maps texture from clothing items to the 3D SCAPE~\cite{SCAPE} body model. Their focus is not photo-realistic mapping, but rather to generate synthetic training data with texture variation to learn discriminative detectors. Their core texture mapping is based on 2D image warping -- unfortunately, the code for this specific part is not available, and therefore comparison is not possible. However, qualitatively, our results look significantly more realistic, and we compare to a very similar baseline based on shape context (SC) matching and Thin Plate Spline (TPS) warping. Furthermore, our approach runs in real time. 
In contrast to all previous work, our method creates textured 3D garments fully automatically, without requiring prior alignment at test-time, which allows virtual try-on and novel view synthesis in real-time.

\section{Method}
\label{sec:fitting}
\begin{figure*}
    \centering
    \begin{overpic}[width=0.9\linewidth]{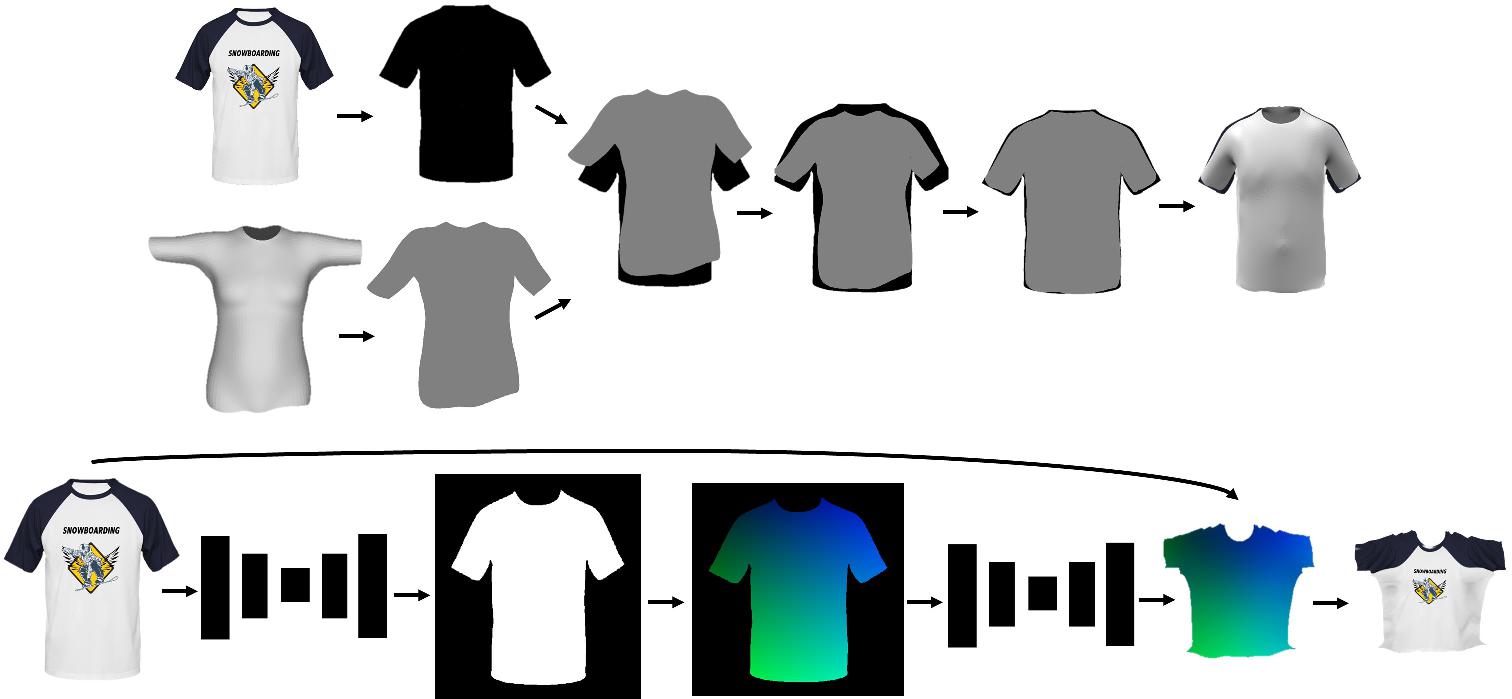} 
    \footnotesize
        \put(-4,20){\rotatebox{90}{\parbox{4cm}{\centering \textcolor{darkgray}{\textsf{ I. Data collection by non-rigid 3D mesh to image alignment}}}}}
        
        \put(-3,-0.5){\rotatebox{90}{\parbox{4cm}{\textcolor{darkgray}{\textsf{ II. Pix2Surf Network}}}}}
    
        \put(14,32.2){Garment}
        \put(29,32.2){Mask}
        
        \put(43.5,24.5){Silhouette matching}
        \put(66.9,24.5){Final match}
        \put(79.8,24.5){Aligned mesh}
        
        \put(3.1,-2){Garment}
        \put(13.7,-2){Segmentation Net}
        \put(33.7,-2){Mask}
        \put(46.2,-2){Image Coordinates}
        \put(64.3,-2){Mapping Net}
        \put(76.1,-2){Correspondences}
        \put(90.5,-2){Final Texture}
    \end{overpic}
	\label{fig:Method}
	\vspace{5mm}
	\centering
	\caption{Overview of method: We build a large paired dataset of 3D garments and online retail store images by non-rigidly aligning 3D garment templates with automatically computed image segmentations (I). This allows us to train our model Pix2Surf (II) to predict correspondences from the UV map of the garment surfaces to image pixel locations. }
\end{figure*}
Our key idea is to learn a \emph{mapping} from images to the UV map of the garment, \emph{without using texture information, but silhouette shape alone}.
We first explain the parametric 3D garment models (Sec.~\ref{subsec:garment}) we use to regularize an automatic mesh to image silhouette fitting procedure (Sec.~\ref{subsec:fitting}).
Since fitting is expensive and error-prone, we learn an efficient neural mapping (Pix2Surf), which transfers the image texture onto the mesh in real time (Sec.~\ref{subsec:learning}).
\subsection{Preliminary: Parametric Garment Model}
\label{subsec:garment}
We leverage publicly available 3D garments templates~\cite{bhatnagar2019mgn} parameterized as displacements from the SMPL body model \cite{smpl2015loper}. 
For every garment category (T-shirt, short pants, long pants), we define a garment template $\posefun^G \in \mathbb{R}^{m\times{3}}$ as a sub-mesh of the SMPL body template $\posefun \in \mathbb{R}^{n\times{3}}$. 
An indicator matrix $\mat{I} \in \mathbb{Z}^{m\times{n}}$ evaluates to $\mathbf{I}_{i,j}=1$ if a garment vertex $i \in \{1 \hdots m\}$ is associated with a body shape vertex $j \in \{1 \hdots n\}$. This correspondence allows representing garments as displacements $\offsets \in \mathbb{R}^{m\times{3}}$ from the unposed SMPL body. Given shape $\shape$ and pose $\pose$, we can articulate a garment using SMPL:
\begin{equation}
\label{eq:garment_shape}
\posefun^G(\shape,\pose,\offsets) = \mat{I}~\posefun(\shape,\pose)+\offsets
\end{equation}
\begin{equation}
\label{eq:garment_pose}
G(\shape,\pose,\mathbf{D}) = \blendfun(\posefun^G(\shape,\pose,\offsets), \jointfun(\shape), \pose, \blendweights),
\end{equation}
with joints $J(\shape)$ and linear blend-skinning $W(\cdot)$ with weights $\mathbf{W}$.
Since in this work we keep $\mathbf{D}$ fixed, we denote the garment model as $G(\pose,\shape)$. After the texture has been transferred to the surface, the geometry can be still be changed with $\mathbf{D}$.
\subsection{Non-Rigid Garment Fitting to Retail Images}
\label{subsec:fitting}
To find a correspondence map between retail images in the web and 3D garments, we could non-rigidly deform its 3D surface to fit the image foreground. 
This is, however, not robust enough as different retailers photograph garments in different poses, backgrounds, and clothing itself varies in shape.  
Hence, we first automatically segment images, and then leverage the parametric garment model $G(\pose,\shape)$ defined in Sec.~\ref{subsec:garment} to regularize the fitting process. 

\subsubsection{Automatic segmentation}
\label{subsubsec:segmentation}
We use an automated version of GrabCut \cite{rother2004grabcut}. Since garments are typically photographed over simple backgrounds, we obtain an approximate foreground mask using thresholding. We then run a closing operation to fill the holes on this mask, and erode it to obtain a prior for ''absolute foreground". The difference region between the mask and its eroded version is marked as ''probable foreground". Analogously, we obtain ''absolute background", and ''probable background" using dilation. Using these prior maps to initialize GrabCut, we obtain accurate silhouettes without manual annotation.
\subsubsection{Garment fitting}

We fit the garment surface to silhouettes in two stages.
In the first stage, we minimize the following objective
\begin{equation}
\mathrm{E_1(\shape, \pose, \trans)} = w_s E_s + w_{\shape} E_{\beta} + w_{\pose} E_{\pose},
\label{eq:fitting}
\end{equation}
w.r.t.\ garment pose, shape and camera translation $\mathbf{t} \in \mathbb{R}^3$.
The objective in Eq.~\eqref{eq:fitting} consists of a silhouette $E_s$, a shape regularization term $E_{\shape}$ and a pose prior term $E_{\pose}$, which we explain in the following. The different terms are balanced using weights $w_*$. 

\noindent
\textbf{Silhouette term:} It is defined as:
\begin{eqnarray}
E_{s}(\shape, \pose, \trans) &=& \Phi(w_i \Psi(I_r(G(\pose,\shape),\trans)) \nonumber \\
&& + w_o \hat{\Psi}(1 - I_r(G(\pose,\shape),\trans))).
\label{eq:silhouette}
\end{eqnarray}
Here, $\Psi$ and $\hat{\Psi}$ are the distance transform, and the inverse distance transform, respectively, of the silhouette image, $\Phi$ is a Gaussian pyramid function, and $I_r(G(\pose,\shape),\trans))$ is the binary garment silhouette image obtained with a differentiable renderer. Consequently, the objective in Eq.~\eqref{eq:silhouette} maximizes overlap between the garment image and the rendered mesh, and penalizes model leackage into the background.

\noindent
\textbf{Shape regularization:} In order to regularize the fitting process in Eq.~\ref{eq:fitting}, we use a Mahalanobis prior 
\begin{equation}
E_{\shape}(\shape) = \shape^T \Sigma^{-1}_{\shape} \shape
\end{equation}
on the shape parameters, where $\Sigma^{-1}_{\shape}$ is the diagonal covariance matrix from the SMPL dataset.

\noindent
\textbf{Pose prior: }The term penalizes deviations of the pose from an A-pose $\pose_A$
\begin{equation}
E_{\pose}(\pose) = \|\pose - \pose_A \|^2
\end{equation}
To minimize Eq.~\eqref{eq:fitting}, we initialize the pose $\pose$ with an A-pose, as this approximates the pose of most garment images on the web.
Additionally, we use scheduling: for shirts we first optimize shape and translation holding pose fixed and optimize all variables jointly afterwards. For pants and shorts the scheduling order is reversed. Stage 1 provides us with a coarse match to the silhouette of the garment, but the final mesh is restricted by the parametric model $G(\pose,\shape, \trans)$. 
To perfectly match silhouette boundaries, we non-rigidly deform free form vertices $\mathbf{G} \in \mathbb{R}^{m\times{3}}$ of a mesh initialized with the optimized garment result of the first stage $G(\pose,\shape)$. Specifically, we optimize a more sophisticated version of Eq.~\ref{eq:fitting}:
\begin{equation}
\mathrm{E_2}(\mathbf{G}, \pose, \shape) =  w'_{s}E'_{s} + w'_{c}E'_{c}  + w'_{l}E'_{l} + w'_{e}E'_e + w'_{b}E'_{b}.
\label{eq:fitting_free}
\end{equation}
$E'_s$ is the same as in Eq.~\eqref{eq:silhouette}, but now we optimize the free form vertices $\mathbf{G}$ instead of the latent parameters of the model $G(\pose,\shape)$.
and $E'_c$, $E'_l$, $E'_e$ and $E'_{b'}$ are coupling, laplacian, edge constraint, and boundary smoothing terms, which we explain next.

\noindent
\textbf{Coupling term:} It penalizes deviations of the free form vertices $\mathbf{G}$ from the parametric garment model $G(\pose,\shape)$:
\begin{equation}
E_c(\mathbf{G},\shape,\pose) = \| \mathbf{G} -  G(\pose,\shape)\|^2
\end{equation}
\textbf{Edge Constraint Term:} The belt or waistline in shorts and pants retail images forms an almost perfect \emph{horizontal line}. We exploit this by matching the top ring (waistline) of pants and shorts 3D meshes to this horizontal line in image space.  
Let  $\mathbf{G}_{i} \in \mathcal{R}$ denote the set of top ring vertices, $\pi(\mathbf{G}_i)_y$ denote the $y$ coordinate of vertex $\mathbf{G}_{i}$ after projection $\pi(\cdot)$ onto the image, and let $y_{max}$ denote the $y$ coordinate of the horizontal line in the image. We incorporate the following penalty: 
\begin{equation}
E_e(\mathbf{G}) = \sum_{\mathbf{G}_{i}  \in {\mathcal{R}}} \| \pi(\mathbf{G}_i)_y - y_{max} \|^2
\end{equation}

\noindent
\textbf{Laplacian Term:}
In order enforce garment smoothness and minimize distortion, we include a Laplacian term~\cite{sorkine2004laplacian}. Given a mesh with adjacency matrix $\mathbf{A}\in \mathbb{R}^{m\times{m}}$ , the graph Laplacian $\mathbf{L} \in \mathbb{R}^{m\times{m}}$ is obtained as $\mathbf{L} = \mathbf{I} - \mathbf{K}^{-1}\mathbf{A}$ where $\mathbf{K}$ is a diagonal matrix such that $\mathbf{K}_{ii}$ stores the number of neighbors of vertex $i$. We minimize the mesh Laplacian:
\begin{equation}
E_l(\mathbf{G}) = || \mathbf{L} \mathbf{G} ||^2_F
\end{equation}
\noindent
\textbf{Boundary Smoothing Term: }To ensure that the boundaries remain smooth, we penalize high second order derivatives along the boundary rings, similar to~\cite{ponsmoll2017clothcap}. \newline
The output of the fitting are 3D garment vertices $ \mathbf{G}$, which together with their faces $\mathbf{F}$ define a deformed mesh $\mathcal{G} = \{\mathbf{G},\mathbf{F}\}$ accurately aligned with image silhouette. 
\subsection{Learning Automatic Texture Transfer}
\label{subsec:learning}
The fitting of the previous section is accurate, but slow, and fails sometimes. Hence, we run the fitting method on internet images, and manually remove the unsuccessful fits. From this data, we train an efficient neural model, referred to as Pix2Surf.
Pix2Surf directly transfers texture from images to the 3D model surface, based on the silhouette shape alone. Next, we explain the key components of Pix2Surf, namely, input output representation, and losses used during training. 
\subsubsection{Pix2Surf: Input and Output Representation}
\label{subsec:input_output}
The curated fits of Sec.~\ref{subsec:fitting} provide dense correspondences from image pixels $(i,j)$ to the 3D garment surface $\mathcal{G} \subset \mathbb{R}^3$. 
Learning a mapping to the surface $\mathcal{G}$ embedded in $\mathbb{R}^3$ is hard, and does not allow leveraging fully 2D convolutional neural neutworks. 
Hence, we compute a 2D UV-map parameterization (Sec.~\ref{subsec:implementation}) of the garment surface, $u: \mathcal{B} \subset \mathbb{R}^2 \mapsto \mathcal{G} \subset \mathbb{R}^3$, 
where $u(\cdot)$ maps $2D$ points $(k,l)$ from the UV space to the surface $\mathcal{G}$ embedded in $\mathbb{R}^3$. In this way, all our targets live in a UV-space; specifically from the fits, we generate: 
RGB texture maps $\mathbf{Y} \in \mathbb{R}^{K\times{L}\times{3}}$ using projective texturing, and \emph{UV correspondence maps}  $\mathbf{C} \in \mathcal{C} \subset \mathbb{R}^{K\times{L}\times{2}}$, which store, at every pixel $(k,l)$ of the (front/back) UV map, the $(i,j)$ coordinates of the corresponding image pixel, that is $\mathbf{C}_{k,l} = (i,j)$.
The input representation for the garment images is a \emph{coordinate mask} $\mathbf{X}\in \mathcal{X} \subset \mathbb{R}^{M\times{N}\times2}$, storing at every image pixel location its own coordinates if the pixel belongs to the foreground $\mathcal{F}$, and $0$ otherwise, 
$\mathbf{X}_{ij} = (i,j)\ \forall  (i,j) \in \mathcal{F}  \,|| \, \mathbf{X}_{ij} = (0,0) \, \forall (i,j) \notin \mathcal{F}$. 
The foreground mask $\mathcal{F}$ is predicted at test time using a \emph{segmentation network} trained using a standard cross-entropy loss--we compute segmentation labels for training using the automatic -- but slow -- GrabCut based method Sec.~\ref{subsubsec:segmentation}.  For the front view of T-shirts, we additionally ask annotators to segment out the back portion of the shirt, which is visible in the front view image. The segmentation network learns to remove this portion ensuring that the back portion is not mapped to the UV-map. 
With this we collect a dataset $\mathcal{D}$ consisting of inputs $\mathbf{X}$, and targets $\{ \mathbf{Y},\mathbf{C}\}$. $\mathcal{D} = \{ \mathbf{X}^i, \{\mathbf{Y}^i, \mathbf{C}^i \} \}_i^N$.
\subsubsection{Pix2Surf: Learning}
We could naively attempt to predict the texture maps $\mathbf{Y}$ directly from images $\mathbf{I}$ using image to image translation, but this is prone to overfit to image textures as we demonstrate in our experiments (Sec.~\ref{subsec:baselines}) .
Instead, we follow a more geometric approach, and learn a \emph{mapping} $f(\mathbf{X};\mathbf{w}):\mathcal{X}\mapsto \mathcal{C}$ from coordinate masks $\mathbf{X}$ to UV correspondence maps $\mathbf{C}$, forcing the network to reason about the input shape. This effectively, learns to predict, for every UV map location $(k,l)$, the corresponding pixel coordinates $( i, j ) $ in the image, $\mathbf{C}_{k,l} = (i,j)$.
Our key insight and assumption is that this smooth mapping depends only on silhouette shape $\mathbf{X}$, and not on texture $\mathbf{I}$. 
During training, we minimize the following loss 
\begin{equation}
L_\mathrm{total} = \lambda_\mathrm{reg} L_\mathrm{reg} + \lambda_\mathrm{perc}L_\mathrm{perc} + \lambda_\mathrm{recon}L_\mathrm{recon}
\end{equation}
over the training set $\mathcal{D}$, where each term in the loss is explained next. \newline
\textbf{Coordinate Regression Loss: } $L_\mathrm{reg}$ evaluates an $L_2(\cdot)$ norm of the difference between the network output $f(\mathbf{X}^i;\mathbf{w})$ and the pseudo ground truths $\mathbf{C}$ obtained using the silhouette fitting algorithm of Sec.~\ref{subsec:fitting}: 
\begin{equation}
L_\mathrm{reg} = \sum_{i=1}^N ||f(\mathbf{X}^i;\mathbf{w}) - \mathbf{C}^i ||^{2}_2 
\end{equation}
\textbf{Reconstruction Loss:} To provide more supervision to the network we use a differentiable sampling kernel ~\cite{jaderberg2015spatial} to infer a texture map directly from the correspondence map. We minimize a dense photometric loss between predicted texture maps and target texture maps $\mathbf{Y}$ obtained with projective texturing (Sec.~\ref{subsec:input_output}):
\begin{equation}
L_\mathrm{recon} = \sum_i^N \sum_{k,l}^{KL} || \mathbf{I}\left[ f_{k,l}^1(\mathbf{X}^i ; \mathbf{w}), f_{k,l}^2(\mathbf{X}^i ; \mathbf{w}) \right] - \mathbf{Y}^i_{k,l}\|_1
\end{equation}
where the original image $\mathbf{I}$ is sampled (using a differentiable kernel) at locations $ (i,j) = (f_{k,l}^1(\mathbf{X}^i ; \mathbf{w}), f_{k,l}^2(\mathbf{X}^i ; \mathbf{w}))$ provided by the predicted correspondence map. \newline
\textbf{Perceptual Loss:} $L_\mathrm{perc}$ is the perceptual loss as defined in~\cite{zhang2018unreasonable} between $\mathbf{I}[f(\mathbf{X};\mathbf{w})] \in \mathbb{R}^{K\times{L}\times{3}}$ (tensor notation) and $\mathbf{Y}$. \newline
Once the network predicts a correspondence map for an input image, we use it and the parallelizable kernel to generate the final image by sampling points from the input image.
\subsubsection{Implementation Details}
\label{subsec:implementation}
We use Adam optimizer for training both networks. For the segmentation network we use a UNet with instance normalization and use color jittering in the input data to improve performance. For Pix2Surf we use a six block ResNet. The choice of normalization and activation functions is same as~\cite{isola2017pix2pix}.\newline
\textbf{Custom UV Map:} Since the artist designed SMPL UV map cuts the garments into different islands (bad for learning a continuous mapping), we use a custom UV map for each garment category. 
We cut the garment surface into front and back and compute the UV map using Blender
This results in two islands (front and back), which makes the image to UV mapping continuous and hence easier to learn.

\section{Experiments}
\label{sec:experiments}
\begin{figure*}
\setlength\tabcolsep{0.0pt}
\begin{tabular}{c c c c c c c c c c c c}
\begin{minipage}{0.083\linewidth}
\includegraphics[width = \linewidth, height = 0.06\paperheight ]{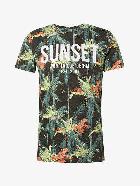}
\\ 
\includegraphics[width = \linewidth, height = 0.06\paperheight]{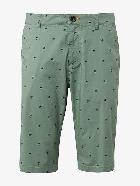}
\end{minipage}
&
\begin{minipage}{0.083\linewidth}
\includegraphics[width = \linewidth, height = 0.06\paperheight ]{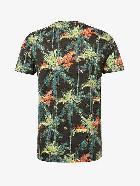}
\\ 
\includegraphics[width = \linewidth, height = 0.06\paperheight]{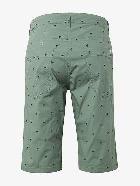}
\end{minipage}
&
\begin{minipage}{0.083\linewidth}
\includegraphics[width = \linewidth, height = 0.12\paperheight]{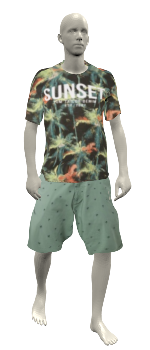}
\end{minipage}
&
\begin{minipage}{0.083\linewidth}
\includegraphics[width = \linewidth, height = 0.12\paperheight]{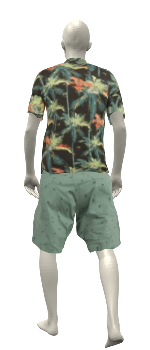}
\end{minipage}
& 

\begin{minipage}{0.083\linewidth}
\includegraphics[width = \linewidth, height = 0.06\paperheight ]{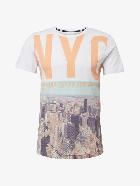}
\\ 
\includegraphics[width = \linewidth, height = 0.06\paperheight]{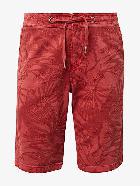}
\end{minipage}
&
\begin{minipage}{0.083\linewidth}
\includegraphics[width = \linewidth, height = 0.06\paperheight ]{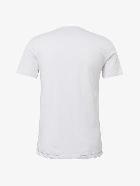}
\\ 
\includegraphics[width = \linewidth, height = 0.06\paperheight]{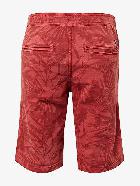}
\end{minipage}
&
\begin{minipage}{0.083\linewidth}
\includegraphics[width = \linewidth, height = 0.12\paperheight]{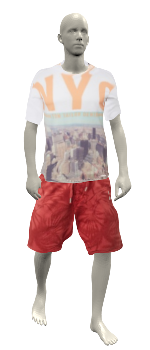}
\end{minipage}
&
\begin{minipage}{0.083\linewidth}
\includegraphics[width = \linewidth, height = 0.12\paperheight]{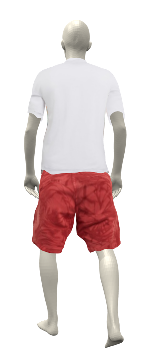}
\end{minipage}

& 

\begin{minipage}{0.083\linewidth}
\includegraphics[width = \linewidth, height = 0.06\paperheight ]{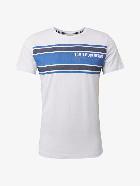}
\\ 
\includegraphics[width = \linewidth, height = 0.06\paperheight]{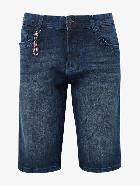}
\end{minipage}
&
\begin{minipage}{0.083\linewidth}
\includegraphics[width = \linewidth, height = 0.06\paperheight ]{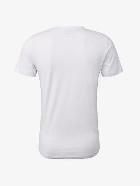}
\\ 
\includegraphics[width = \linewidth, height = 0.06\paperheight]{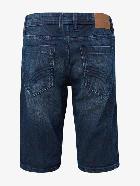}
\end{minipage}
&
\begin{minipage}{0.083\linewidth}
\includegraphics[width = \linewidth, height = 0.12\paperheight]{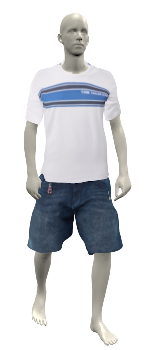}
\end{minipage}
&
\begin{minipage}{0.083\linewidth}
\includegraphics[width = \linewidth, height = 0.12\paperheight]{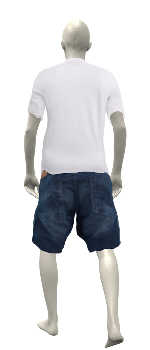}
\end{minipage}
\end{tabular}
\begin{tabular}{c c c c c c c c c c c c}
\begin{minipage}{0.083\linewidth}
\includegraphics[width = \linewidth, height = 0.06\paperheight ]{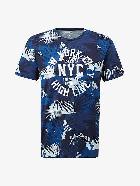}
\\ 
\includegraphics[width = \linewidth, height = 0.06\paperheight]{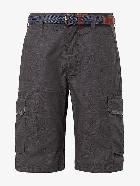}
\end{minipage}
&
\begin{minipage}{0.083\linewidth}
\includegraphics[width = \linewidth, height = 0.06\paperheight ]{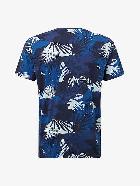}
\\ 
\includegraphics[width = \linewidth, height = 0.06\paperheight]{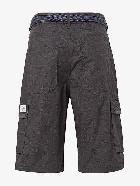}
\end{minipage}
&
\begin{minipage}{0.083\linewidth}
\includegraphics[width = \linewidth, height = 0.12\paperheight]{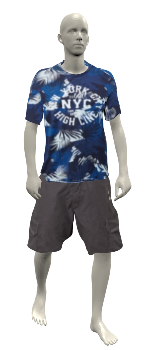}
\end{minipage}
&
\begin{minipage}{0.083\linewidth}
\includegraphics[width = \linewidth, height = 0.12\paperheight]{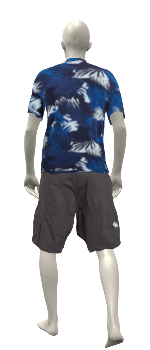}
\end{minipage}
& 

\begin{minipage}{0.083\linewidth}
\includegraphics[width = \linewidth, height = 0.06\paperheight ]{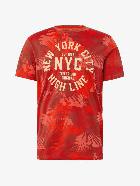}
\\ 
\includegraphics[width = \linewidth, height = 0.06\paperheight]{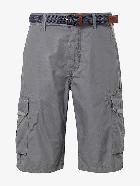}
\end{minipage}
&
\begin{minipage}{0.083\linewidth}
\includegraphics[width = \linewidth, height = 0.06\paperheight ]{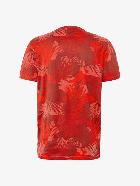}
\\ 
\includegraphics[width = \linewidth, height = 0.06\paperheight]{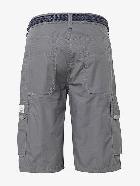}
\end{minipage}
&
\begin{minipage}{0.083\linewidth}
\includegraphics[width = \linewidth, height = 0.12\paperheight]{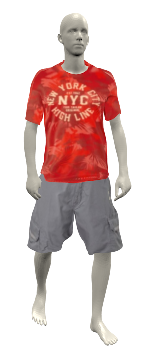}
\end{minipage}
&
\begin{minipage}{0.083\linewidth}
\includegraphics[width = \linewidth, height = 0.12\paperheight]{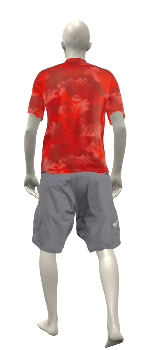}
\end{minipage}
& 
\begin{minipage}{0.083\linewidth}
\includegraphics[width = \linewidth, height = 0.06\paperheight ]{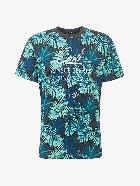}
\\ 
\includegraphics[width = \linewidth, height = 0.06\paperheight]{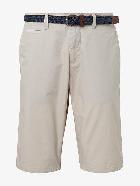}
\end{minipage}
&
\begin{minipage}{0.083\linewidth}
\includegraphics[width = \linewidth, height = 0.06\paperheight ]{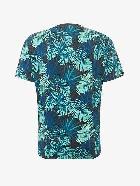}
\\ 
\includegraphics[width = \linewidth, height = 0.06\paperheight]{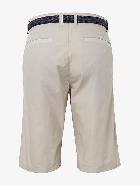}
\end{minipage}
&
\begin{minipage}{0.083\linewidth}
\includegraphics[width = \linewidth, height = 0.12\paperheight]{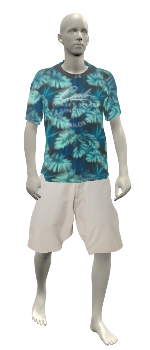}
\end{minipage}
&
\begin{minipage}{0.083\linewidth}
\includegraphics[width = \linewidth, height = 0.12\paperheight]{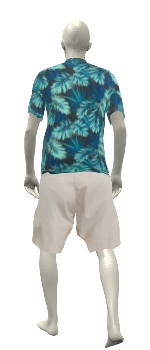}
\end{minipage}
\end{tabular}

\begin{tabular}{c c c c c c c c c c c c}
\begin{minipage}{0.083\linewidth}
\includegraphics[width = \linewidth, height = 0.06\paperheight ]{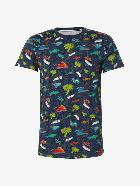}
\\ 
\includegraphics[width = \linewidth, height = 0.06\paperheight]{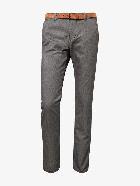}
\end{minipage}
&
\begin{minipage}{0.083\linewidth}
\includegraphics[width = \linewidth, height = 0.06\paperheight ]{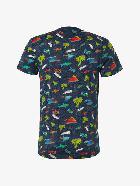}
\\ 
\includegraphics[width = \linewidth, height = 0.06\paperheight]{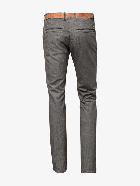}
\end{minipage}
&
\begin{minipage}{0.083\linewidth}
\includegraphics[width = \linewidth, height = 0.12\paperheight]{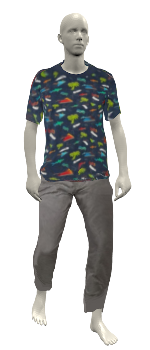}
\end{minipage}
&
\begin{minipage}{0.083\linewidth}
\includegraphics[width = \linewidth, height = 0.12\paperheight]{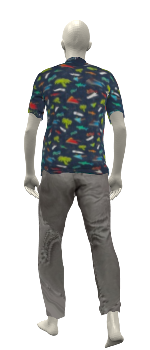}
\end{minipage}
&

\begin{minipage}{0.083\linewidth}
\includegraphics[width = \linewidth, height = 0.06\paperheight ]{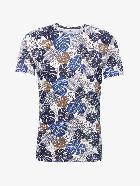}
\\ 
\includegraphics[width = \linewidth, height = 0.06\paperheight]{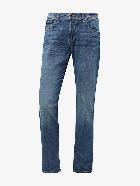}
\end{minipage}
&
\begin{minipage}{0.083\linewidth}
\includegraphics[width = \linewidth, height = 0.06\paperheight ]{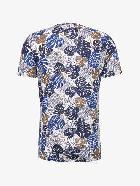}
\\ 
\includegraphics[width = \linewidth, height = 0.06\paperheight]{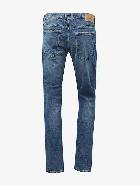}
\end{minipage}
&
\begin{minipage}{0.083\linewidth}
\includegraphics[width = \linewidth, height = 0.12\paperheight]{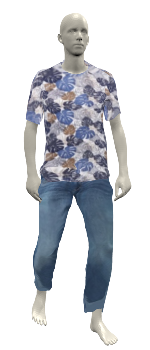}
\end{minipage}
&
\begin{minipage}{0.083\linewidth}
\includegraphics[width = \linewidth, height = 0.12\paperheight]{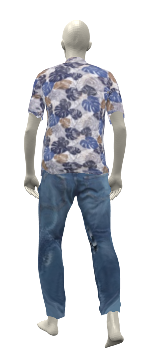}
\end{minipage}

& 
\begin{minipage}{0.083\linewidth}
\includegraphics[width = \linewidth, height = 0.06\paperheight ]{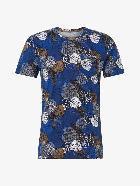}
\\ 
\includegraphics[width = \linewidth, height = 0.06\paperheight]{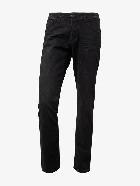}
\end{minipage}
&
\begin{minipage}{0.083\linewidth}
\includegraphics[width = \linewidth, height = 0.06\paperheight ]{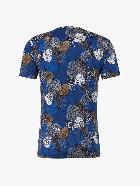}
\\ 
\includegraphics[width = \linewidth, height = 0.06\paperheight]{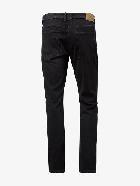}
\end{minipage}
&
\begin{minipage}{0.083\linewidth}
\includegraphics[width = \linewidth, height = 0.12\paperheight]{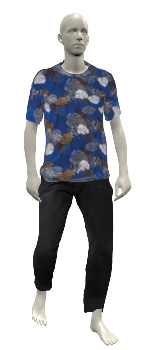}
\end{minipage}
&
\begin{minipage}{0.083\linewidth}
\includegraphics[width = \linewidth, height = 0.12\paperheight]{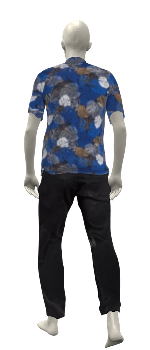}
\end{minipage}


\end{tabular}


\begin{tabular}{c c c c c c c c c c c c}
\begin{minipage}{0.083\linewidth}
\includegraphics[width = \linewidth, height = 0.06\paperheight ]{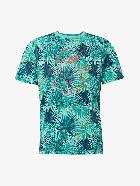}
\\ 
\includegraphics[width = \linewidth, height = 0.06\paperheight]{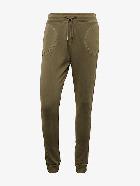}
\end{minipage}
&
\begin{minipage}{0.083\linewidth}
\includegraphics[width = \linewidth, height = 0.06\paperheight ]{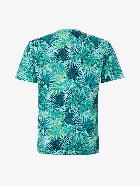}
\\ 
\includegraphics[width = \linewidth, height = 0.06\paperheight]{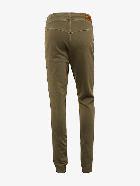}
\end{minipage}
&
\begin{minipage}{0.083\linewidth}
\includegraphics[width = \linewidth, height = 0.12\paperheight]{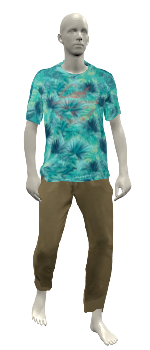}
\end{minipage}
&
\begin{minipage}{0.083\linewidth}
\includegraphics[width = \linewidth, height = 0.12\paperheight]{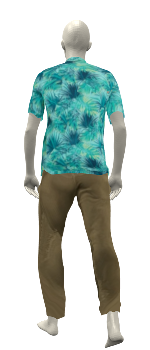}
\end{minipage}
&

\begin{minipage}{0.083\linewidth}
\includegraphics[width = \linewidth, height = 0.06\paperheight ]{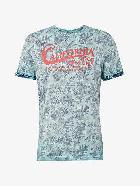}
\\ 
\includegraphics[width = \linewidth, height = 0.06\paperheight]{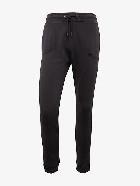}
\end{minipage}
&
\begin{minipage}{0.083\linewidth}
\includegraphics[width = \linewidth, height = 0.06\paperheight ]{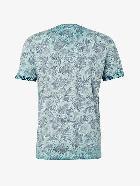}
\\ 
\includegraphics[width = \linewidth, height = 0.06\paperheight]{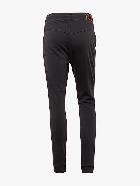}
\end{minipage}
&
\begin{minipage}{0.083\linewidth}
\includegraphics[width = \linewidth, height = 0.12\paperheight]{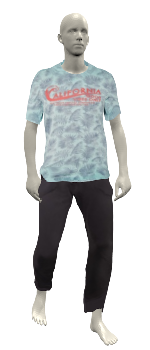}
\end{minipage}
&
\begin{minipage}{0.083\linewidth}
\includegraphics[width = \linewidth, height = 0.12\paperheight]{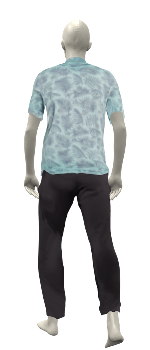}
\end{minipage}

& 
\begin{minipage}{0.083\linewidth}
\includegraphics[width = \linewidth, height = 0.06\paperheight ]{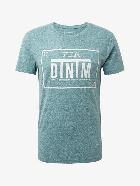}
\\ 
\includegraphics[width = \linewidth, height = 0.06\paperheight]{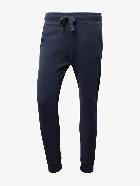}
\end{minipage}
&
\begin{minipage}{0.083\linewidth}
\includegraphics[width = \linewidth, height = 0.06\paperheight ]{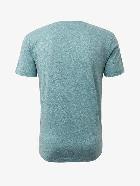}
\\ 
\includegraphics[width = \linewidth, height = 0.06\paperheight]{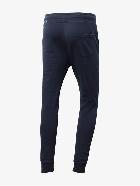}
\end{minipage}
&
\begin{minipage}{0.083\linewidth}
\includegraphics[width = \linewidth, height = 0.12\paperheight]{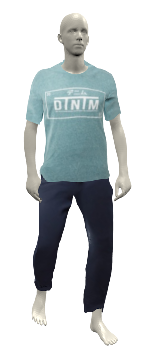}
\end{minipage}
&
\begin{minipage}{0.083\linewidth}
\includegraphics[width = \linewidth, height = 0.12\paperheight]{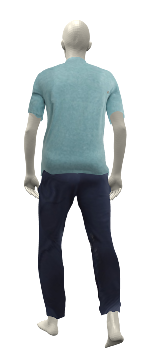}
\end{minipage}


\end{tabular}

\caption{Textured garments obtained using our method: From the online retail store images (left), we create textures for three different garment categories (T-shirt, pants, shorts). We use the textured garments to virtually dress SMPL (right).}
\label{fig:results}
\end{figure*}
Since we are solving a problem for which there is no ground-truth data, we evaluate our proposed method qualitatively and with a user study. 
We show results for three different garment types (T-shirts, shorts and long-pants), see Fig.~\ref{fig:results}. Notice that the texture is nicely mapped without transferring background. Notice also how the 3D textures are complete without \emph{holes and seams}. 
We compare our method against the popular Shape-context with Thin Plate Spline (TPS) matching baseline, an image-to-image translation which operates on pixels directly, and image based virtual try on methods.
\subsection{Dataset}
To train our models, we create datasets of garment images by scraping the websites of clothing stores -- specifically the websites of Zalando \cite{zalando}, Tom Tailor \cite{tom-tailor} and Jack and Jones \cite{jack-jones}.
The back view of clothing images is scarcely available on the internet, so we were unable to create a dataset of back view images large enough for training mapping networks. We leverage the fact that the distribution of garment silhouettes is similar for both front and back views, and either augment the dataset used for training the mapping from the back view image to the back UV map by combining front view images with back view images or using \emph{only} front view images.  \newline
We create a datasets consisting of 2267 front images of T-shirts.  The networks pertaining to the back are trained using a dataset of 2267 images of which 964 are back view images and the rest are front images. The front shorts dataset has 2277 items. We use the same dataset to train the networks pertaining to the back views. For pants we collect a dataset of 3410 front views. We create a dataset for back views by horizontally flipping the front view images and their corresponding silhouettes. The back dataset has 3211 items. The failure of the optimization based registration method explains the discrepancy between the number of items in the front and back views. Exploiting these front-back symmetries turns out to work well in practice, and allows to compensate for the unavailability of back view images.
\subsection{Shape Context Matching and Image-to-Image Translation Baselines}
\label{subsec:baselines}
\begin{figure*}
\setlength\tabcolsep{0.0pt}
\begin{tabular}{c c c c c c c c c c c c}
\begin{minipage}{0.083\linewidth}
\includegraphics[width = \linewidth, height = 0.06\paperheight ]{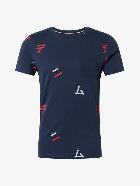}
\\ 
\includegraphics[width = \linewidth, height = 0.06\paperheight]{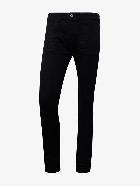}
\end{minipage}
&
\begin{minipage}{0.083\linewidth}
\includegraphics[width = \linewidth, height = 0.12\paperheight]{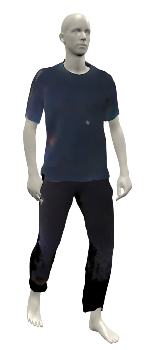}
\end{minipage}
&
\begin{minipage}{0.083\linewidth}
\begin{overpic}[width = \linewidth, height = 0.12\paperheight]{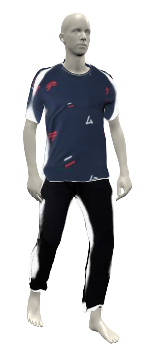}
\put(0,0){\color{red}\linethickness{0.1mm}
\polygon(7,64)(7,80)(15,80)(15,64)
\polygon(30,65)(30,80)(35,80)(35,65)
\polygon(11,50)(15,50)(15,63)(11,63)
\polygon(10,20)(10,43)(14,43)(14,20)
\polygon(23,43)(27,43)(27,15)(23,15)

}

\end{overpic}
\end{minipage}
&
\begin{minipage}{0.083\linewidth}
\includegraphics[width = \linewidth, height = 0.12\paperheight]{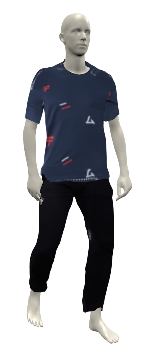}
\end{minipage}

& 
\begin{minipage}{0.083\linewidth}
\includegraphics[width = \linewidth, height = 0.06\paperheight ]{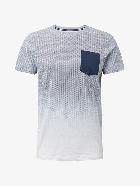}
\\ 
\includegraphics[width = \linewidth, height = 0.06\paperheight]{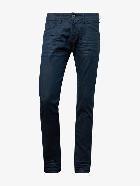}
\end{minipage}
&
\begin{minipage}{0.083\linewidth}
\includegraphics[width = \linewidth, height = 0.12\paperheight]{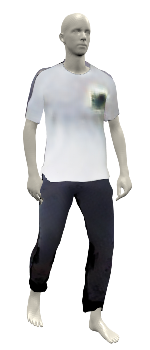}
\end{minipage}
&
\begin{minipage}{0.083\linewidth}

\begin{overpic}[width = \linewidth, height = 0.12\paperheight]{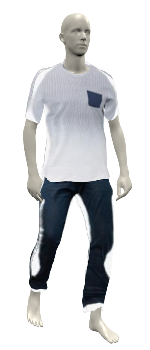}
\put(0,0){\color{red}\linethickness{0.1mm}
\polygon(7,64)(7,80)(15,80)(15,64)
\polygon(30,65)(30,80)(35,80)(35,65)
\polygon(11,50)(15,50)(15,63)(11,63)
\polygon(10,20)(10,43)(14,43)(14,20)
\polygon(23,43)(27,43)(27,15)(23,15)
}
\end{overpic}
\end{minipage}
&
\begin{minipage}{0.083\linewidth}
\includegraphics[width = \linewidth, height = 0.12\paperheight]{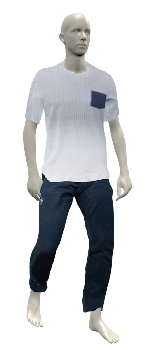}
\end{minipage}
\begin{minipage}{0.083\linewidth}
\includegraphics[width = \linewidth, height = 0.06\paperheight ]{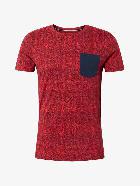}
\\ 
\includegraphics[width = \linewidth, height = 0.06\paperheight]{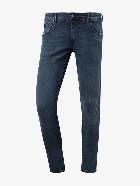}
\end{minipage}
&
\begin{minipage}{0.083\linewidth}
\includegraphics[width = \linewidth, height = 0.12\paperheight]{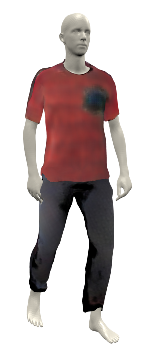}

\end{minipage}
&
\begin{minipage}{0.083\linewidth}
\begin{overpic}[width = \linewidth, height = 0.12\paperheight]{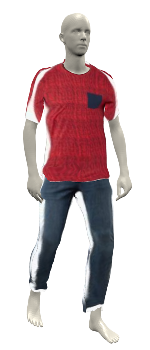}
\put(0,0){\color{red}\linethickness{0.1mm}
\polygon(7,64)(7,80)(15,80)(15,64)
\polygon(30,65)(30,80)(35,80)(35,65)
\polygon(11,50)(15,50)(15,63)(11,63)
\polygon(10,20)(10,43)(14,43)(14,20)
\polygon(23,43)(27,43)(27,15)(23,15)

}

\end{overpic}
\end{minipage}
&
\begin{minipage}{0.083\linewidth}
\includegraphics[width = \linewidth, height = 0.12\paperheight]{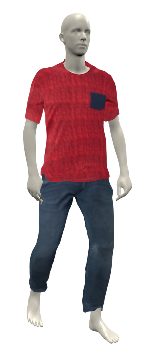}
\end{minipage}
\end{tabular}

\begin{tabular}{c c c c c c c c c c c c}
\begin{minipage}{0.083\linewidth}
\includegraphics[width = \linewidth, height = 0.06\paperheight ]{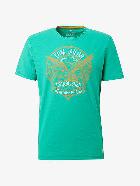}
\\ 
\includegraphics[width = \linewidth, height = 0.06\paperheight]{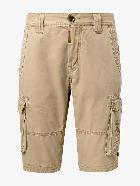}
\end{minipage}
&
\begin{minipage}{0.083\linewidth}
\includegraphics[width = \linewidth, height = 0.12\paperheight]{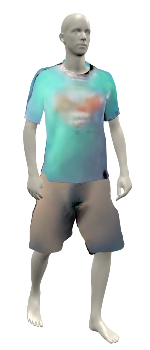}
\end{minipage}
&
\begin{minipage}{0.083\linewidth}
\begin{overpic}[ width = \linewidth, height = 0.12\paperheight]{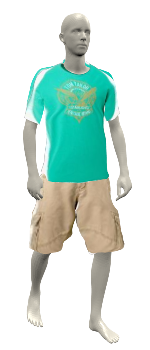}
\put(0,0){\color{red}\linethickness{0.1mm}
\polygon(7,64)(7,80)(15,80)(15,64)
\polygon(30,65)(30,80)(35,80)(35,65)
\polygon(11,50)(15,50)(15,63)(11,63)
\polygon(8,30)(8,45)(12,45)(12,30)
}

\end{overpic}
\end{minipage}
&
\begin{minipage}{0.083\linewidth}
\includegraphics[width = \linewidth, height = 0.12\paperheight]{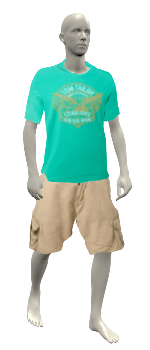}
\end{minipage}

& 
\begin{minipage}{0.083\linewidth}
\includegraphics[width = \linewidth, height = 0.06\paperheight ]{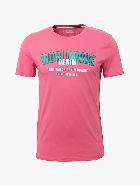}
\\ 
\includegraphics[width = \linewidth, height = 0.06\paperheight]{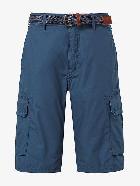}
\end{minipage}
&
\begin{minipage}{0.083\linewidth}
\includegraphics[width = \linewidth, height = 0.12\paperheight]{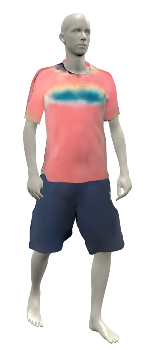}
\end{minipage}
&
\begin{minipage}{0.083\linewidth}
\begin{overpic}[ width = \linewidth, height = 0.12\paperheight]{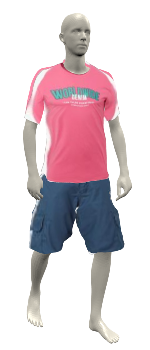}
\put(0,0){\color{red}\linethickness{0.1mm}
\polygon(7,64)(7,80)(15,80)(15,64)
\polygon(30,65)(30,80)(35,80)(35,65)
\polygon(11,50)(15,50)(15,63)(11,63)
\polygon(8,30)(8,45)(12,45)(12,30)
}

\end{overpic}
\end{minipage}
&
\begin{minipage}{0.083\linewidth}
\includegraphics[width = \linewidth, height = 0.12\paperheight]{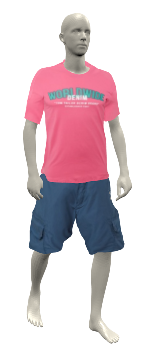}
\end{minipage}

\begin{minipage}{0.083\linewidth}
\includegraphics[width = \linewidth, height = 0.06\paperheight ]{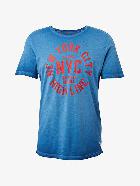}
\\ 
\includegraphics[width = \linewidth, height = 0.06\paperheight]{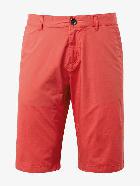}
\end{minipage}
&
\begin{minipage}{0.083\linewidth}
\includegraphics[width = \linewidth, height = 0.12\paperheight]{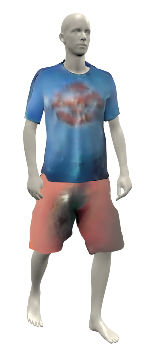}
\end{minipage}
&
\begin{minipage}{0.083\linewidth}
\begin{overpic}[ width = \linewidth, height = 0.12\paperheight]{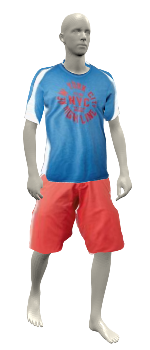}
\put(0,0){\color{red}\linethickness{0.1mm}
\polygon(7,64)(7,80)(15,80)(15,64)
\polygon(30,65)(30,80)(35,80)(35,65)
\polygon(11,50)(15,50)(15,63)(11,63)
\polygon(8,30)(8,45)(12,45)(12,30)
}

\end{overpic}
\end{minipage}
&
\begin{minipage}{0.083\linewidth}
\includegraphics[width = \linewidth, height = 0.12\paperheight]{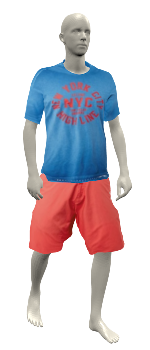}
\end{minipage}
\end{tabular}
\caption{We compare our results against textures obtained via SC matching and an image-to-image translation baseline. SC matching produces textures with artefacts and Pix2Pix produces blurry results. Left: Pix2Pix, Middle: SC matching, Right: Pix2Surf \vspace{-2.2 em}}
\label{fig:comp_p2p_warp}
\end{figure*}

We compare our method with shape context (SC) matching plus TPS and image-to-image translation baselines. We implement the shape context matching baseline as follows. We first render the garment mesh to obtain a silhouette. To make the baseline as strong as possible, we first pose the garment in the same pose as the image. We then warp the garment image to match the rendered garment. We then lift the texture from the warped image onto the mesh using projective texturing. Our experiments demonstrate that SC matching and TPS~\cite{belongie2002shape} is not powerful enough to precisely match the contours of the two shapes so the final texture always has artifacts along the sides. See  Fig.~\ref{fig:comp_p2p_warp}. \newline
We also compare Pix2Surf to a Pix2Pix~\cite{isola2017pix2pix} baseline, which learns to produce a texture map from an image. To train Pix2Pix, we use the same texture maps we use to train Pix2Surf. Unlike Pix2Surf, Pix2Pix produces blurry results and does not preserve details of the input texture. See Fig.~\ref{fig:comp_p2p_warp}.  \newline
For further evaluation, we have conducted a \emph{user study} with 30 participants. We created 20 videos containing images of reference clothing textures and two rotating textured 3D avatars ``wearing" the reference clothing. One is textured using a baseline method and the other using Pix2Surf. We ask participants to choose the better looking avatar. In 100\% of all comparisons, the avatar textured with Pix2Surf was preferred over the one textured using the baselines. 
\subsection{Towards Photo-Realistic Rendering}
\label{subsec:photoreal}
\begin{figure}
\setlength\tabcolsep{0.0pt}
\begin{tabular}{c c c c c c}
\begin{minipage}{0.20\linewidth}
\includegraphics[width = \linewidth, height = 0.08\paperheight ]{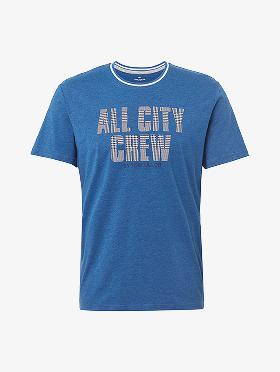}
\end{minipage}
&
\begin{minipage}{0.20\linewidth}
\includegraphics[width = \linewidth, height = 0.08\paperheight]{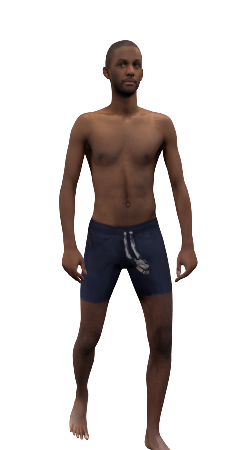}
\end{minipage}
&
\begin{minipage}{0.20\linewidth}
\includegraphics[width = \linewidth, height = 0.08\paperheight]{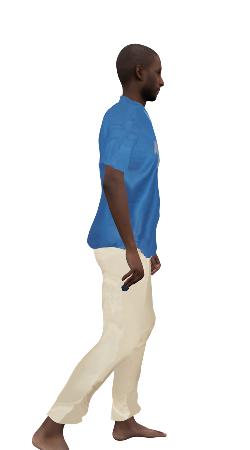}
\end{minipage}
&
\begin{minipage}{0.20\linewidth}
\includegraphics[width = \linewidth, height = 0.08\paperheight]{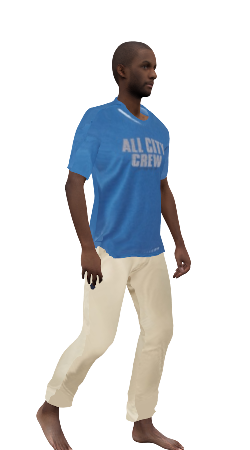}
\end{minipage}
&
\begin{minipage}{0.20\linewidth}
\includegraphics[width = \linewidth, height = 0.08\paperheight]{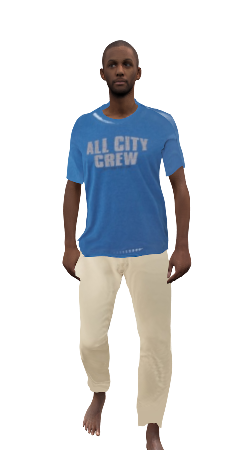}
\end{minipage}
\end{tabular}
\begin{tabular}{c c c c c c}
\begin{minipage}{0.20\linewidth}
 \includegraphics[width = \linewidth, height = 0.08\paperheight]{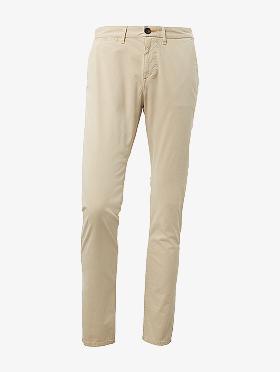}
\end{minipage}
&
\begin{minipage}{0.20\linewidth}
\includegraphics[width = \linewidth, height = 0.08\paperheight]{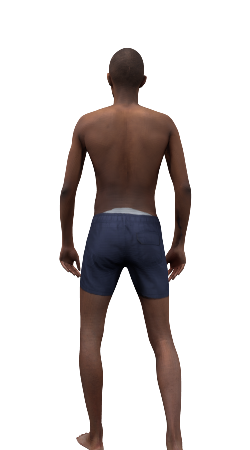}
\end{minipage}
&
\begin{minipage}{0.20\linewidth}
\includegraphics[width = \linewidth, height = 0.08\paperheight]{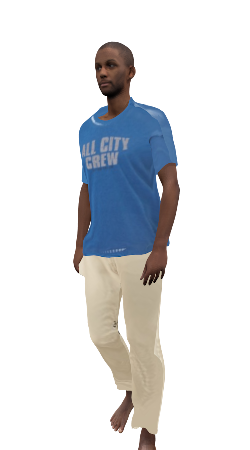}
\end{minipage}
&
\begin{minipage}{0.20\linewidth}
\includegraphics[width = \linewidth, height = 0.08\paperheight]{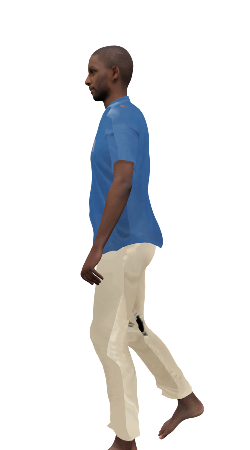}
\end{minipage}
&
\begin{minipage}{0.20\linewidth}
\includegraphics[width = \linewidth, height = 0.08\paperheight]{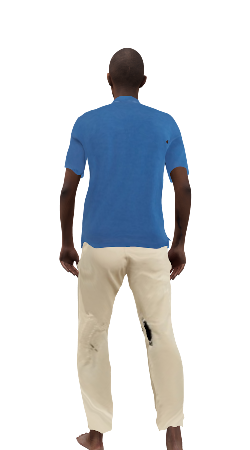}
\end{minipage}
\end{tabular}
\vspace{2mm}
\caption{Photo-realistic virtual try-on. Given the garment images and a 3D avatar with texture, we show our automatically textured garments on top. \vspace{-2.2 em}}
\label{fig:photoreal}
\end{figure}

Provided with the texture map of a 3D avatar without clothes, Pix2Surf allows to map the texture atop the avatar from potentially an infinite number of clothing items scraped from online websites, \emph{automatically} and in \emph{real time}. See Fig. \ref{fig:photoreal}
Using these photo-realistic renderings, we are able to compare our method with Image based virtual try-on method such as VITON \cite{han2017viton} and CP-VTON \cite{wang2018toward}. Since our method requires a 3D avatar as input, to be able to compare, we first render a photo-realistic image using our method, and use VITON to change the upper garment of the rendering. 
Fig.~\ref{fig:VITON} shows that VITON and CP-VTON work reasonably well for some poses, but fail badly for extreme poses while our method does not. We note that these methods are related but conceptually different compared to our method, so a direct fair comparison is not possible. These methods are also not explicitly trained to handle such poses, but the figure illustrates a general limitation of image based virtual try-on approaches to generalize to novel view-points and poses. In stark contrast, we only need to generalize to silhouette shape variation, and once the texture is mapped to the 3D avatar, novel poses and viewpoints are trivially generated applying transformation in 3D.
\begin{figure}
\setlength\tabcolsep{0.0pt}
\begin{tabular}{c c c c c}
\begin{minipage}{0.20\linewidth}
\includegraphics[width = \linewidth, height = 0.08\paperheight ]{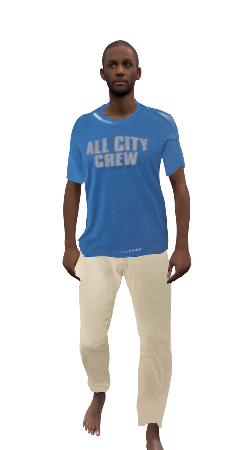}
\\ 
\includegraphics[width = \linewidth, height = 0.08\paperheight]{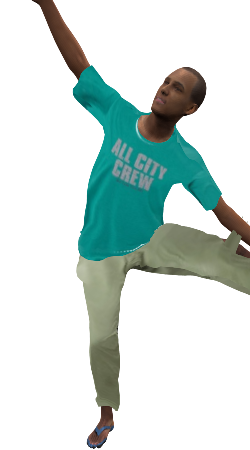} 
\\
\includegraphics[width = \linewidth, height = 0.08\paperheight]{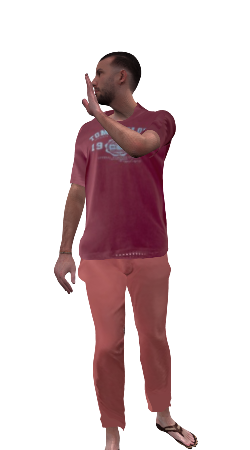}
\\
\includegraphics[width = \linewidth, height = 0.08\paperheight]{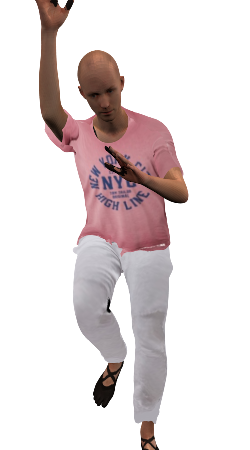}
\\
\includegraphics[width = \linewidth, height = 0.08\paperheight]{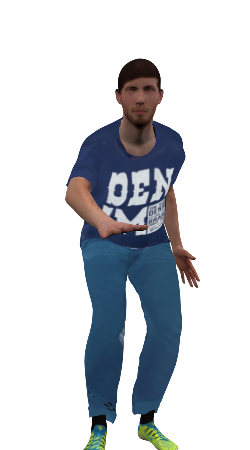}
\end{minipage}
&
\begin{minipage}{0.20\linewidth}
\includegraphics[width = \linewidth, height = 0.08\paperheight ]{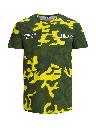}
\\ 
\includegraphics[width = \linewidth, height = 0.08\paperheight]{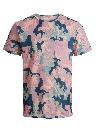} 
\\
\includegraphics[width = \linewidth, height = 0.08\paperheight]{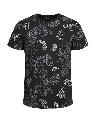}
\\
\includegraphics[width = \linewidth, height = 0.08\paperheight]{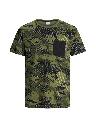}
\\
\includegraphics[width = \linewidth, height = 0.08\paperheight]{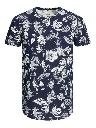}
\end{minipage}

&

\begin{minipage}{0.20\linewidth}
\includegraphics[width = \linewidth, height = 0.08\paperheight ]{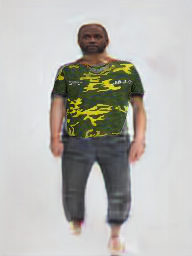}
\\ 
\includegraphics[width = \linewidth, height = 0.08\paperheight]{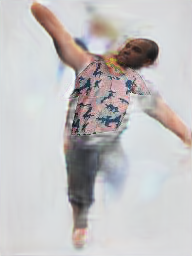}
\\ 
\includegraphics[width = \linewidth, height = 0.08\paperheight]{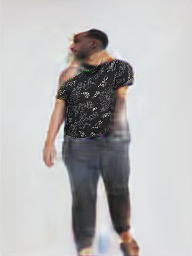}
\\ 
\includegraphics[width = \linewidth, height = 0.08\paperheight]{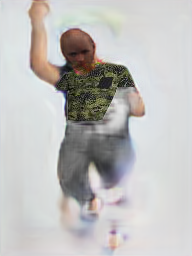}
\\ 
\includegraphics[width = \linewidth, height = 0.08\paperheight]{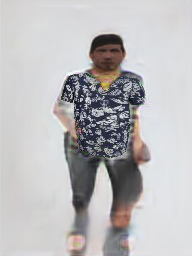}
\end{minipage}

&

\begin{minipage}{0.20\linewidth}
\includegraphics[width = \linewidth, height = 0.08\paperheight ]{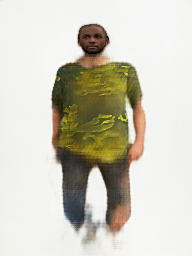}
\\ 
\includegraphics[width = \linewidth, height = 0.08\paperheight]{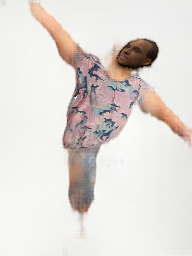}
\\ 
\includegraphics[width = \linewidth, height = 0.08\paperheight]{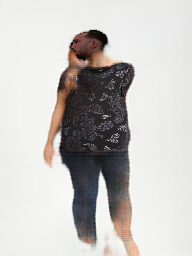}
\\ 
\includegraphics[width = \linewidth, height = 0.08\paperheight]{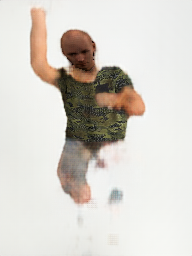}
\\ 
\includegraphics[width = \linewidth, height = 0.08\paperheight]{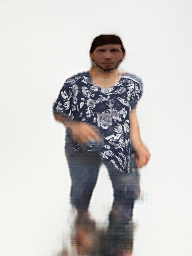}
\end{minipage}

&

\begin{minipage}{0.20\linewidth}
\includegraphics[width = \linewidth, height = 0.08\paperheight ]{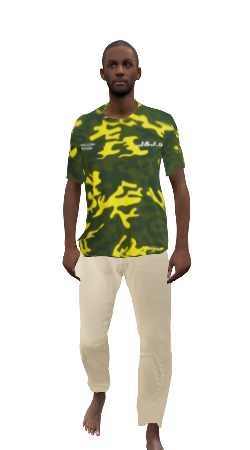}
\\ 
\includegraphics[width = \linewidth, height = 0.08\paperheight]{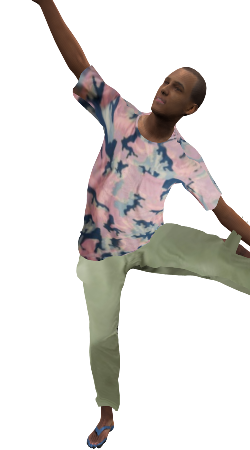}
\\ 
\includegraphics[width = \linewidth, height = 0.08\paperheight]{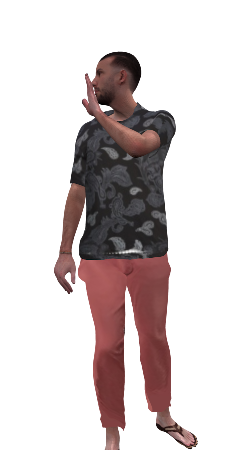}
\\ 
\includegraphics[width = \linewidth, height = 0.08\paperheight]{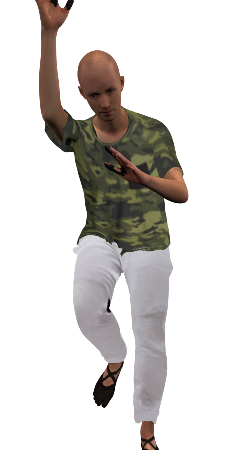}
\\ 
\includegraphics[width = \linewidth, height = 0.08\paperheight]{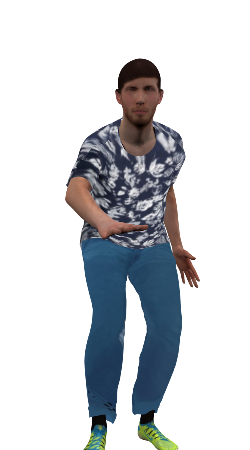}
\end{minipage}
\end{tabular}
\begin{tabu} to \linewidth {X[c]X[c]X[c]X[c]X[c]}
 Input image & Target Cloth & CP-VTON & VITON & Ours \\
\end{tabu}
\vspace{2mm}
\caption{Comparison with VITON and CP-VTON \vspace{-2.2 em}}
\label{fig:VITON}
\end{figure}

\section{Conclusions and Future Work}
\label{sec:conclusion}
We have presented a simple yet effective model that learns to transfer textures from web images of clothing to 3D garments worn by virtual humans. 
Our experiments show that our non-linear optimization method is accurate enough to compute a training set of clothing images aligned with 3D mesh projections, 
from which we learn a direct mapping with a neural model (Pix2Surf). 
While the optimization method takes up ten minutes to converge, Pix2Surf runs in real time, which is crucial for many applications such as virtual try-on. 
Our key idea is to learn the correspondence map from image pixels to a 2D UV parameterization of the surface, \emph{based on silhouette shape alone instead of texture}, which makes the model invariant to the highly varying textures of clothing and consequently generalize better.    
We show Pix2Surf performs significantly better than classical approaches such as 2D TPS warping (while being orders of magnitude faster), and direct image-to-image translation approaches. 

We believe our model represents an important step towards learning a generative model of textures directly in 3D.
We plan to address this in the future since it is lacking in current models like SMPL~\cite{smpl2015loper}. We focused on texture, and assume garment geometry is given, but we think it should be possible to infer geometry from images. Since clothing is laid out on a flat surface on web photographs, geometry inference will require modelling how 3D garments deform when they are flattened out on a surface. 

\begin{minipage}{\columnwidth}
    \footnotesize
\noindent
\textbf{Acknowledgments: }We thank Bharat Lal Bhatnagar and Garvita Tiwari for in-depth  discussions and assistance  in  our  data preparation. This  work  is  partly  funded  by  the  Deutsche  Forschungsgemeinschaft  (DFG) - 409792180 (Emmy Noether Programme, project:  Real Virtual Humans).
\end{minipage}

{\small
\bibliographystyle{ieee_fullname}
\bibliography{egbib}
}
\appendix
\section{Non-Rigid Garment Fitting}
In all our experiments, we use the neutral gendered SMPL model. For each garment class we only use a subset of angles in the input axis-angle representation of pose as our optimization variables. For shirts, we use only the pose parameters corresponding to the shoulder joints - numbered $13,14,16,17$ in the canonical ordering. For shorts, joint corresponding to the hip joint - joints $1$ and $2$ are used. Joints $1,2,3,4$ which correspond to the hip and knee joints are used while registering pants.  \newline
The weights used in the optimization based registration are given below. For registration the front of T-shirts, we set $w_{\shape} = 3.0, w_s = 100, w_{\pose} = 1.0, w'_{s} = 2.0, w'_c = 0.3, w'_{b} = 2.0, w'_l = 30.0, w'_e = 0$. The same weights when registration the back are:  $w_{\shape} = 5.0, w_s = 100, w_{\pose} = 0, w'_{s} = 2.0, w'_c = 0.3, w'_{b} = 2.0, w'_l = 30.0, w'_e = 0$. \newline
For the front view of shorts, we set $w_{\shape} = 4.0, w_s = 100$ $ w_{\pose} = 30.0$ in the first stage when shape is not being optimized and $ w_{\pose} = 20$ when it is. $w'_{s} = 100, w'_c = 4.0, w'_{b} = 2.0, w'_l = 45.0, w'_e = 0.001$. For the back $w_{\shape} = 3.0, w_s = 160.0$ $ w_{\pose} = 40.0$ when shape is not an optimization variable and $ w_{\pose} = 30$ when it is. $w'_{s} = 100, w'_c = 3.8, w'_{b} = 6.0, w'_l = 75.0, w'_e = 0.01$. \newline
For both front and back of pants, we set $w_{\shape} = 4.0, w_s = 100$ $ w_{\pose} = 40.0$ in the stage when just the pose is optimized $ w_{\pose} = 5$ when both shape and pose are optimization variables. $w'_{s} = 100, w'_c = 0.5, w'_{b} = 2.0, w'_l = 45.0, w'_e = 0.1$. 
\section{Learning Automatic Texture Transfer}
For all our experiments, we use an Adam optimizer with $\beta1 = 0.5$ and $\beta2 = 0.999$. We use a constant learning rate of $0.0001$. All mapping networks are trained for $200$ epochs and all segmentation networks are trained for $20$ epochs. During the training of the segmentation networks we use heavy colour jittering to prevent the network from overfitting to the textures in the training data. The weights are set as $\lambda_{reg} = \lambda_{recon} = \lambda_{perc} = 1$ \newline
\section{Baseline Implementation}
For the SC-matching TPS based baseline, all parameters are set to their default values. The pix2pix baseline is implemented with the weight $\lambda_{gan} = 1$. We observed that increasing the weight of the adversarial loss worsens performance. 
\section{Custom UV Maps}
\begin{figure}
    \centering
    \includegraphics[width = \linewidth]{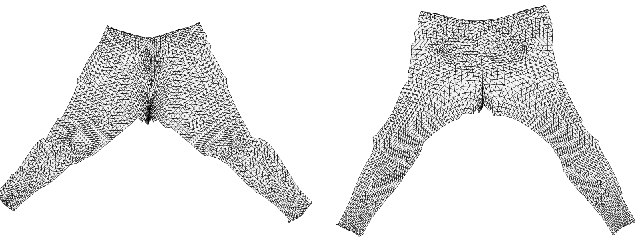}
    \includegraphics[width = \linewidth]{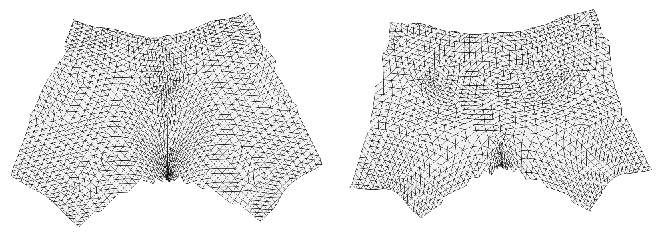}
    \includegraphics[width = \linewidth]{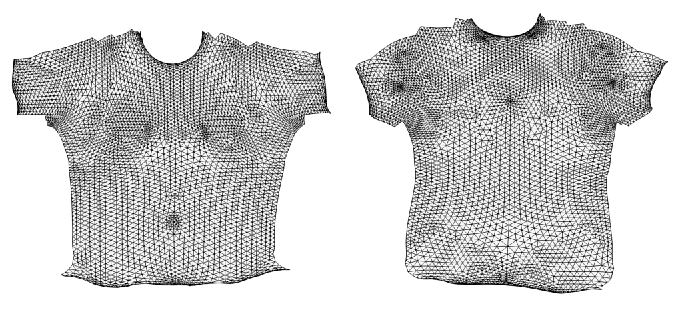}
    \caption{Custom UV maps}
    \label{fig:uv_maps}
\end{figure}

All the UV maps used are created in blender. We obtained best performance when the shape of the UV maps are similar to the shapes of the input garments. The exact UV maps used can be seen in Fig. \ref{fig:uv_maps} 
\section{Results}
\begin{figure*}
\setlength\tabcolsep{0.0pt}
\begin{tabular}{c c c c c c c c c c c c}
\begin{minipage}{0.083\linewidth}
\includegraphics[width = \linewidth, height = 0.06\paperheight ]{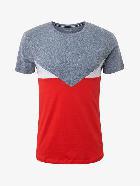}
\\ 
\includegraphics[width = \linewidth, height = 0.06\paperheight]{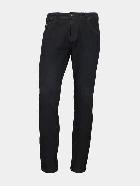}
\end{minipage}
&
\begin{minipage}{0.083\linewidth}
\includegraphics[width = \linewidth, height = 0.06\paperheight ]{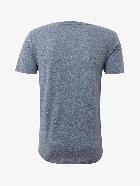}
\\ 
\includegraphics[width = \linewidth, height = 0.06\paperheight]{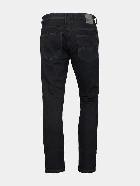}
\end{minipage}
&
\begin{minipage}{0.083\linewidth}
\includegraphics[width = \linewidth, height = 0.12\paperheight]{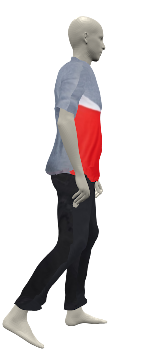}
\end{minipage}
&
\begin{minipage}{0.083\linewidth}
\includegraphics[width = \linewidth, height = 0.12\paperheight]{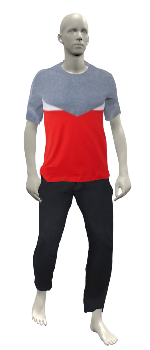}
\end{minipage}
&
\begin{minipage}{0.083\linewidth}
\includegraphics[width = \linewidth, height = 0.12\paperheight]{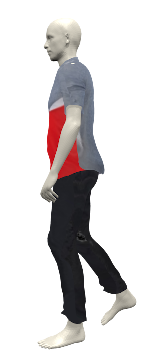}
\end{minipage}
&
\begin{minipage}{0.083\linewidth}
\includegraphics[width = \linewidth, height = 0.12\paperheight]{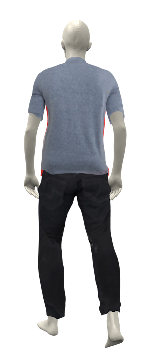}
\end{minipage}
\begin{minipage}{0.083\linewidth}
\includegraphics[width = \linewidth, height = 0.06\paperheight ]{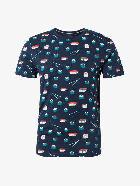}
\\ 
\includegraphics[width = \linewidth, height = 0.06\paperheight]{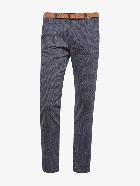}
\end{minipage}
&
\begin{minipage}{0.083\linewidth}
\includegraphics[width = \linewidth, height = 0.06\paperheight ]{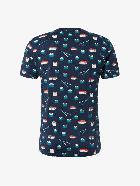}
\\ 
\includegraphics[width = \linewidth, height = 0.06\paperheight]{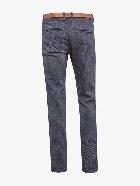}
\end{minipage}
&
\begin{minipage}{0.083\linewidth}
\includegraphics[width = \linewidth, height = 0.12\paperheight]{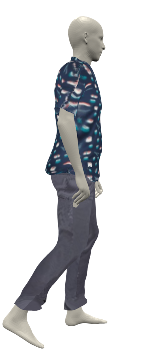}
\end{minipage}
&
\begin{minipage}{0.083\linewidth}
\includegraphics[width = \linewidth, height = 0.12\paperheight]{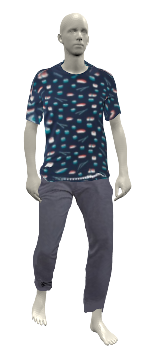}
\end{minipage}
&
\begin{minipage}{0.083\linewidth}
\includegraphics[width = \linewidth, height = 0.12\paperheight]{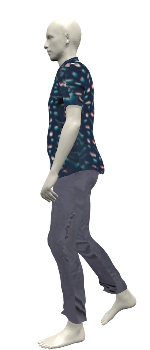}
\end{minipage}
&
\begin{minipage}{0.083\linewidth}
\includegraphics[width = \linewidth, height = 0.12\paperheight]{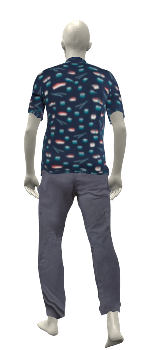}
\end{minipage}
\end{tabular}

\begin{tabular}{c c c c c c c c c c c c}
\begin{minipage}{0.083\linewidth}
\includegraphics[width = \linewidth, height = 0.06\paperheight ]{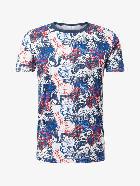}
\\ 
\includegraphics[width = \linewidth, height = 0.06\paperheight]{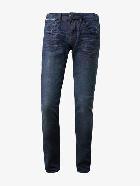}
\end{minipage}
&
\begin{minipage}{0.083\linewidth}
\includegraphics[width = \linewidth, height = 0.06\paperheight ]{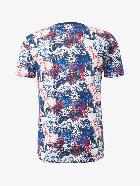}
\\ 
\includegraphics[width = \linewidth, height = 0.06\paperheight]{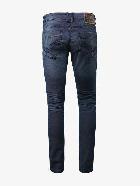}
\end{minipage}
&
\begin{minipage}{0.083\linewidth}
\includegraphics[width = \linewidth, height = 0.12\paperheight]{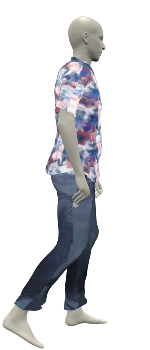}
\end{minipage}
&
\begin{minipage}{0.083\linewidth}
\includegraphics[width = \linewidth, height = 0.12\paperheight]{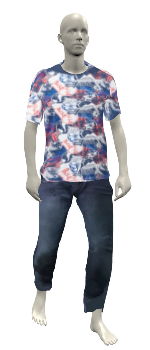}
\end{minipage}
&
\begin{minipage}{0.083\linewidth}
\includegraphics[width = \linewidth, height = 0.12\paperheight]{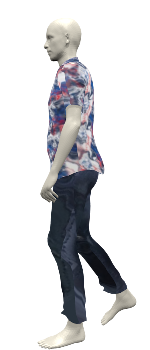}
\end{minipage}
&
\begin{minipage}{0.083\linewidth}
\includegraphics[width = \linewidth, height = 0.12\paperheight]{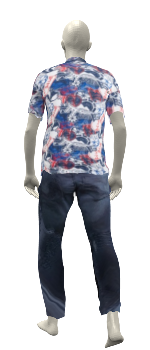}
\end{minipage}
\begin{minipage}{0.083\linewidth}
\includegraphics[width = \linewidth, height = 0.06\paperheight ]{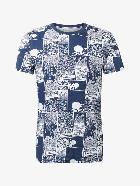}
\\ 
\includegraphics[width = \linewidth, height = 0.06\paperheight]{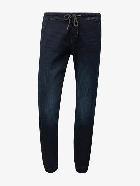}
\end{minipage}
&
\begin{minipage}{0.083\linewidth}
\includegraphics[width = \linewidth, height = 0.06\paperheight ]{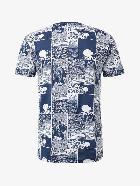}
\\ 
\includegraphics[width = \linewidth, height = 0.06\paperheight]{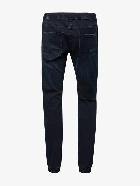}
\end{minipage}
&
\begin{minipage}{0.083\linewidth}
\includegraphics[width = \linewidth, height = 0.12\paperheight]{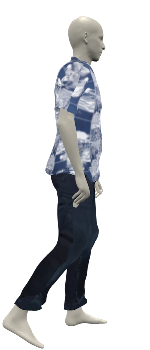}
\end{minipage}
&
\begin{minipage}{0.083\linewidth}
\includegraphics[width = \linewidth, height = 0.12\paperheight]{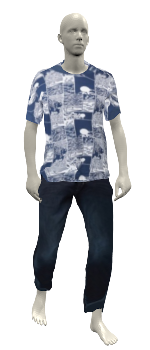}
\end{minipage}
&
\begin{minipage}{0.083\linewidth}
\includegraphics[width = \linewidth, height = 0.12\paperheight]{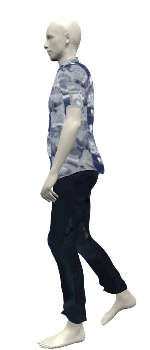}
\end{minipage}
&
\begin{minipage}{0.083\linewidth}
\includegraphics[width = \linewidth, height = 0.12\paperheight]{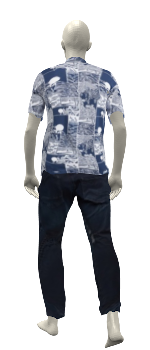}
\end{minipage}
\end{tabular}

\begin{tabular}{c c c c c c c c c c c c}
\begin{minipage}{0.083\linewidth}
\includegraphics[width = \linewidth, height = 0.06\paperheight ]{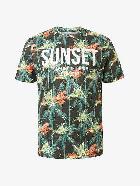}
\\ 
\includegraphics[width = \linewidth, height = 0.06\paperheight]{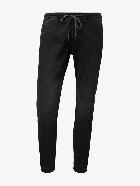}
\end{minipage}
&
\begin{minipage}{0.083\linewidth}
\includegraphics[width = \linewidth, height = 0.06\paperheight ]{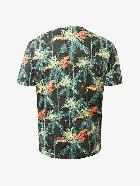}
\\ 
\includegraphics[width = \linewidth, height = 0.06\paperheight]{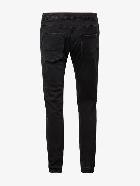}
\end{minipage}
&
\begin{minipage}{0.083\linewidth}
\includegraphics[width = \linewidth, height = 0.12\paperheight]{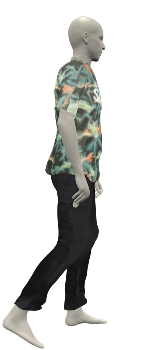}
\end{minipage}
&
\begin{minipage}{0.083\linewidth}
\includegraphics[width = \linewidth, height = 0.12\paperheight]{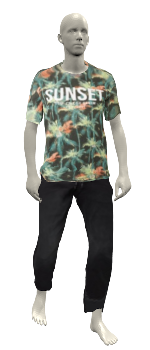}
\end{minipage}
&
\begin{minipage}{0.083\linewidth}
\includegraphics[width = \linewidth, height = 0.12\paperheight]{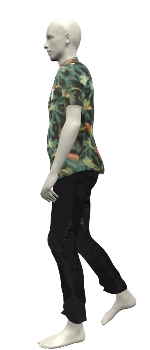}
\end{minipage}
&
\begin{minipage}{0.083\linewidth}
\includegraphics[width = \linewidth, height = 0.12\paperheight]{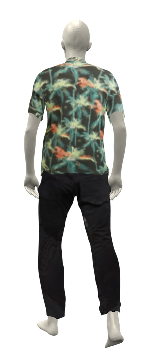}
\end{minipage}
\begin{minipage}{0.083\linewidth}
\includegraphics[width = \linewidth, height = 0.06\paperheight ]{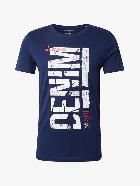}
\\ 
\includegraphics[width = \linewidth, height = 0.06\paperheight]{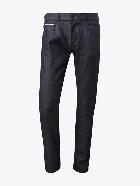}
\end{minipage}
&
\begin{minipage}{0.083\linewidth}
\includegraphics[width = \linewidth, height = 0.06\paperheight ]{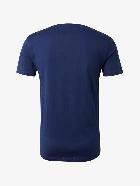}
\\ 
\includegraphics[width = \linewidth, height = 0.06\paperheight]{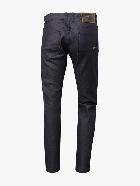}
\end{minipage}
&
\begin{minipage}{0.083\linewidth}
\includegraphics[width = \linewidth, height = 0.12\paperheight]{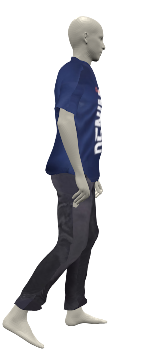}
\end{minipage}
&
\begin{minipage}{0.083\linewidth}
\includegraphics[width = \linewidth, height = 0.12\paperheight]{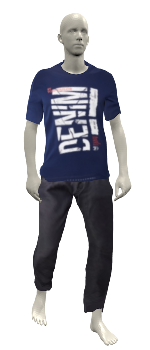}
\end{minipage}
&
\begin{minipage}{0.083\linewidth}
\includegraphics[width = \linewidth, height = 0.12\paperheight]{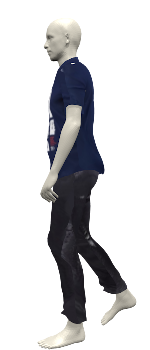}
\end{minipage}
&
\begin{minipage}{0.083\linewidth}
\includegraphics[width = \linewidth, height = 0.12\paperheight]{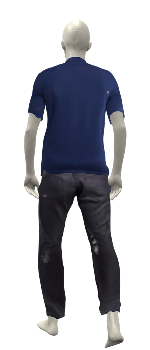}
\end{minipage}
\end{tabular}

\begin{tabular}{c c c c c c c c c c c c}
\begin{minipage}{0.083\linewidth}
\includegraphics[width = \linewidth, height = 0.06\paperheight ]{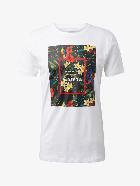}
\\ 
\includegraphics[width = \linewidth, height = 0.06\paperheight]{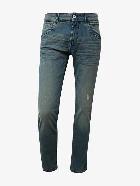}
\end{minipage}
&
\begin{minipage}{0.083\linewidth}
\includegraphics[width = \linewidth, height = 0.06\paperheight ]{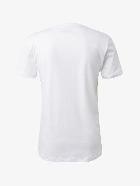}
\\ 
\includegraphics[width = \linewidth, height = 0.06\paperheight]{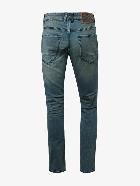}
\end{minipage}
&
\begin{minipage}{0.083\linewidth}
\includegraphics[width = \linewidth, height = 0.12\paperheight]{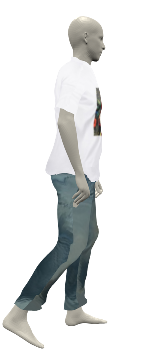}
\end{minipage}
&
\begin{minipage}{0.083\linewidth}
\includegraphics[width = \linewidth, height = 0.12\paperheight]{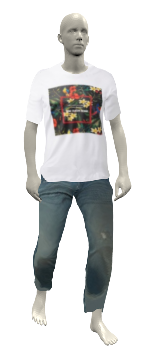}
\end{minipage}
&
\begin{minipage}{0.083\linewidth}
\includegraphics[width = \linewidth, height = 0.12\paperheight]{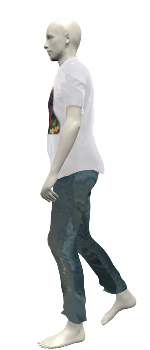}
\end{minipage}
&
\begin{minipage}{0.083\linewidth}
\includegraphics[width = \linewidth, height = 0.12\paperheight]{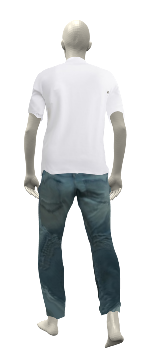}
\end{minipage}
\begin{minipage}{0.083\linewidth}
\includegraphics[width = \linewidth, height = 0.06\paperheight ]{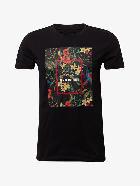}
\\ 
\includegraphics[width = \linewidth, height = 0.06\paperheight]{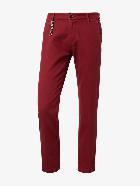}
\end{minipage}
&
\begin{minipage}{0.083\linewidth}
\includegraphics[width = \linewidth, height = 0.06\paperheight ]{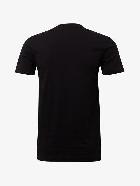}
\\ 
\includegraphics[width = \linewidth, height = 0.06\paperheight]{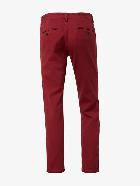}
\end{minipage}
&
\begin{minipage}{0.083\linewidth}
\includegraphics[width = \linewidth, height = 0.12\paperheight]{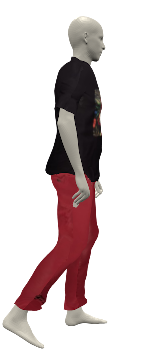}
\end{minipage}
&
\begin{minipage}{0.083\linewidth}
\includegraphics[width = \linewidth, height = 0.12\paperheight]{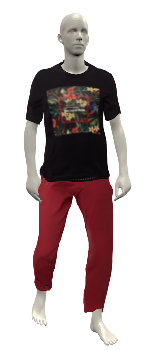}
\end{minipage}
&
\begin{minipage}{0.083\linewidth}
\includegraphics[width = \linewidth, height = 0.12\paperheight]{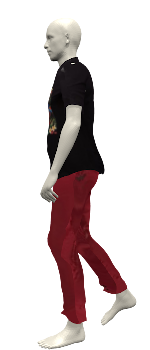}
\end{minipage}
&
\begin{minipage}{0.083\linewidth}
\includegraphics[width = \linewidth, height = 0.12\paperheight]{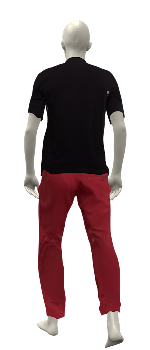}
\end{minipage}
\end{tabular}

\begin{tabular}{c c c c c c c c c c c c}
\begin{minipage}{0.083\linewidth}
\includegraphics[width = \linewidth, height = 0.06\paperheight ]{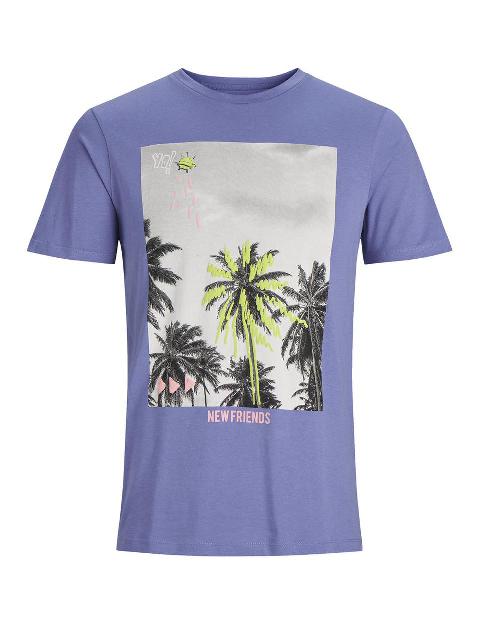}
\\ 
\includegraphics[width = \linewidth, height = 0.06\paperheight]{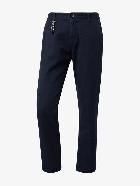}
\end{minipage}
&
\begin{minipage}{0.083\linewidth}
\includegraphics[width = \linewidth, height = 0.06\paperheight ]{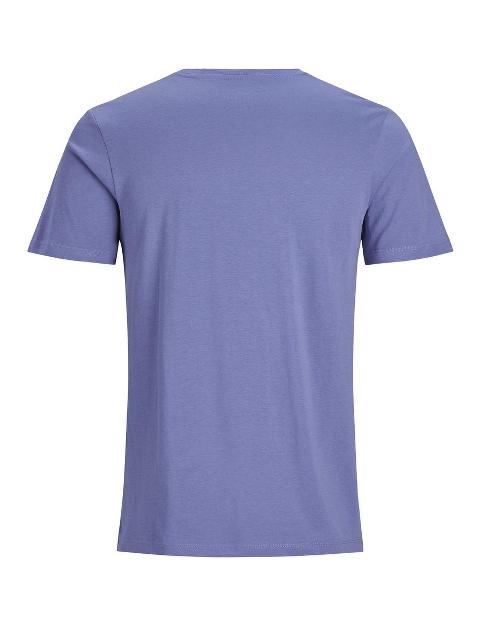}
\\ 
\includegraphics[width = \linewidth, height = 0.06\paperheight]{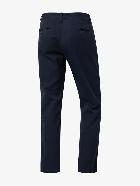}
\end{minipage}
&
\begin{minipage}{0.083\linewidth}
\includegraphics[width = \linewidth, height = 0.12\paperheight]{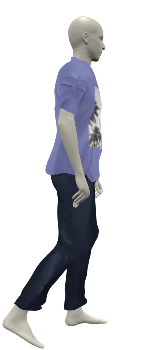}
\end{minipage}
&
\begin{minipage}{0.083\linewidth}
\includegraphics[width = \linewidth, height = 0.12\paperheight]{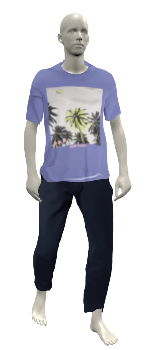}
\end{minipage}
&
\begin{minipage}{0.083\linewidth}
\includegraphics[width = \linewidth, height = 0.12\paperheight]{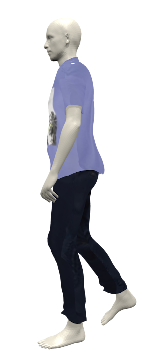}
\end{minipage}
&
\begin{minipage}{0.083\linewidth}
\includegraphics[width = \linewidth, height = 0.12\paperheight]{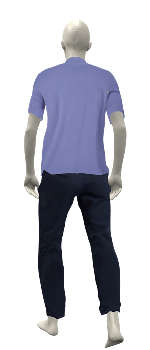}
\end{minipage}
\begin{minipage}{0.083\linewidth}
\includegraphics[width = \linewidth, height = 0.06\paperheight ]{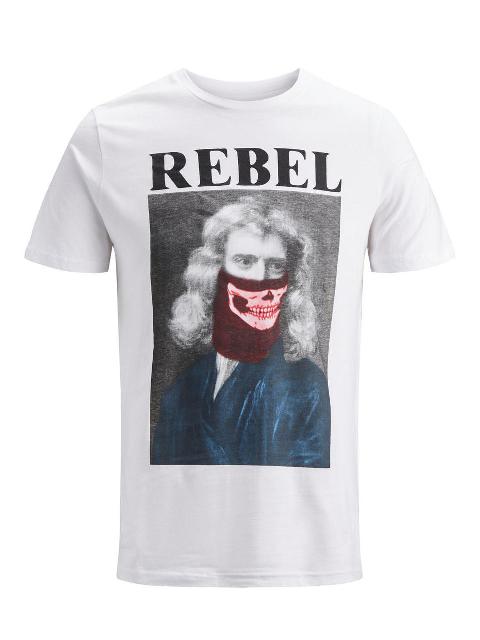}
\\ 
\includegraphics[width = \linewidth, height = 0.06\paperheight]{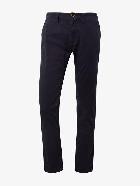}
\end{minipage}
&
\begin{minipage}{0.083\linewidth}
\includegraphics[width = \linewidth, height = 0.06\paperheight ]{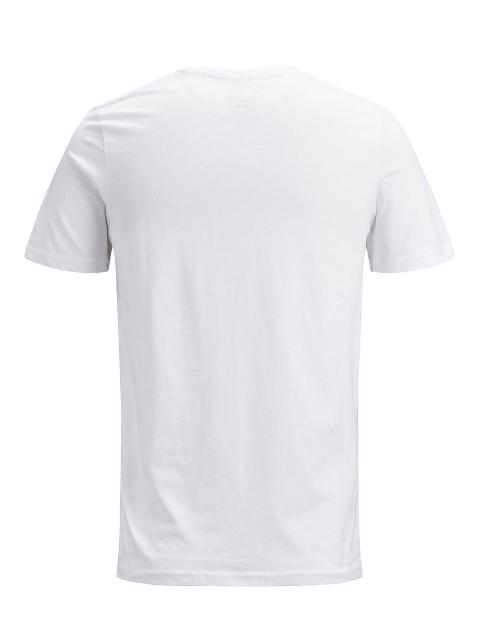}
\\ 
\includegraphics[width = \linewidth, height = 0.06\paperheight]{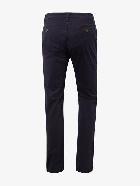}
\end{minipage}
&
\begin{minipage}{0.083\linewidth}
\includegraphics[width = \linewidth, height = 0.12\paperheight]{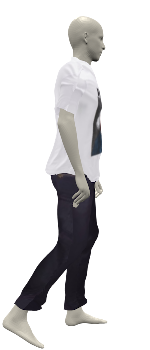}
\end{minipage}
&
\begin{minipage}{0.083\linewidth}
\includegraphics[width = \linewidth, height = 0.12\paperheight]{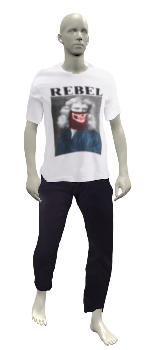}
\end{minipage}
&
\begin{minipage}{0.083\linewidth}
\includegraphics[width = \linewidth, height = 0.12\paperheight]{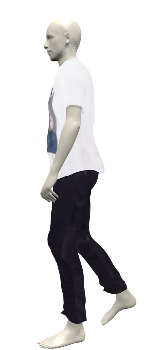}
\end{minipage}
&
\begin{minipage}{0.083\linewidth}
\includegraphics[width = \linewidth, height = 0.12\paperheight]{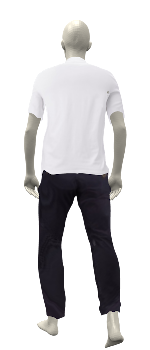}
\end{minipage}
\end{tabular}

\begin{tabular}{c c c c c c c c c c c c}
\begin{minipage}{0.083\linewidth}
\includegraphics[width = \linewidth, height = 0.06\paperheight ]{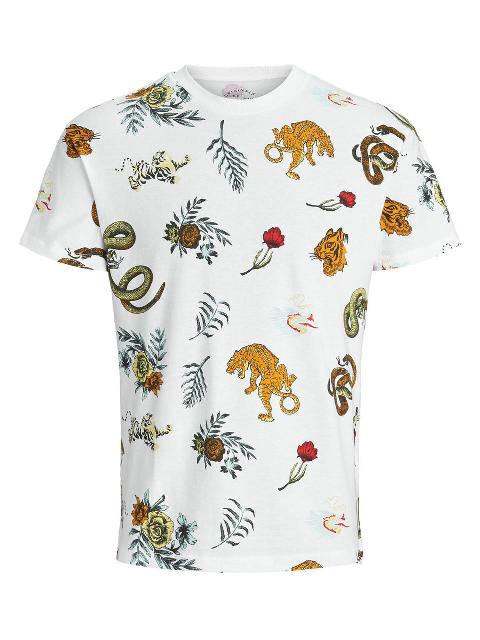}
\\ 
\includegraphics[width = \linewidth, height = 0.06\paperheight]{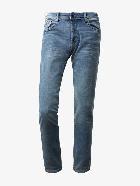}
\end{minipage}
&
\begin{minipage}{0.083\linewidth}
\includegraphics[width = \linewidth, height = 0.06\paperheight ]{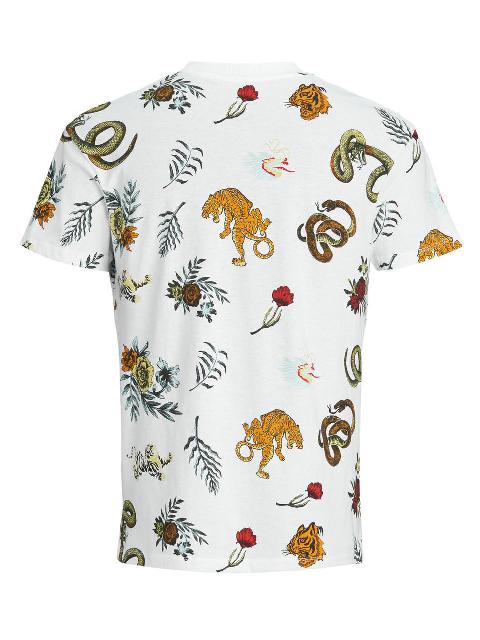}
\\ 
\includegraphics[width = \linewidth, height = 0.06\paperheight]{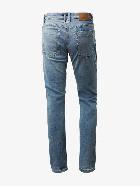}
\end{minipage}
&
\begin{minipage}{0.083\linewidth}
\includegraphics[width = \linewidth, height = 0.12\paperheight]{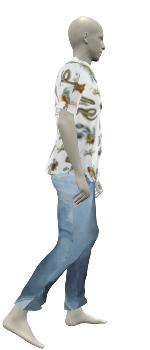}
\end{minipage}
&
\begin{minipage}{0.083\linewidth}
\includegraphics[width = \linewidth, height = 0.12\paperheight]{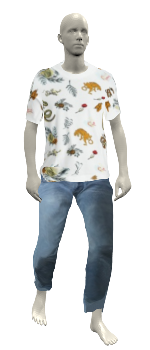}
\end{minipage}
&
\begin{minipage}{0.083\linewidth}
\includegraphics[width = \linewidth, height = 0.12\paperheight]{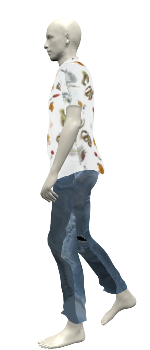}
\end{minipage}
&
\begin{minipage}{0.083\linewidth}
\includegraphics[width = \linewidth, height = 0.12\paperheight]{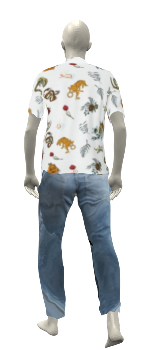}
\end{minipage}
\begin{minipage}{0.083\linewidth}
\includegraphics[width = \linewidth, height = 0.06\paperheight ]{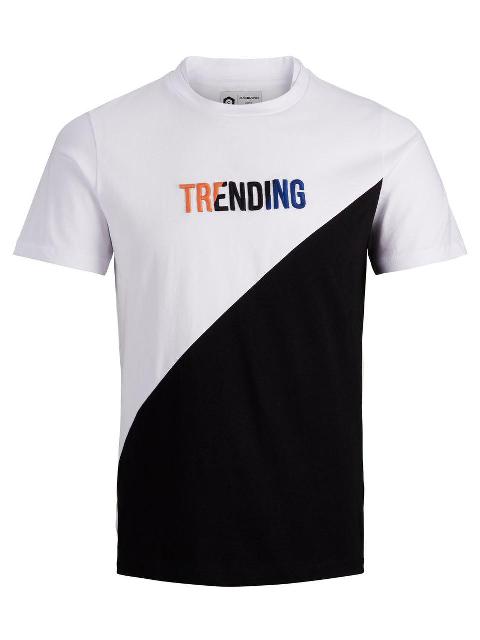}
\\ 
\includegraphics[width = \linewidth, height = 0.06\paperheight]{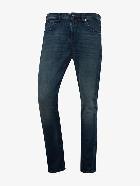}
\end{minipage}
&
\begin{minipage}{0.083\linewidth}
\includegraphics[width = \linewidth, height = 0.06\paperheight ]{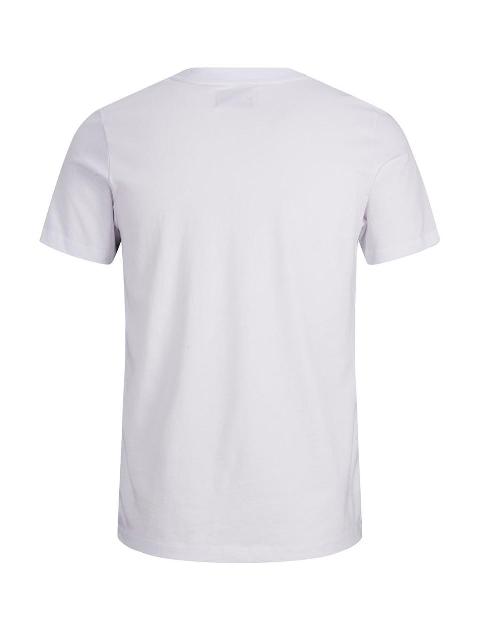}
\\ 
\includegraphics[width = \linewidth, height = 0.06\paperheight]{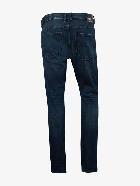}
\end{minipage}
&
\begin{minipage}{0.083\linewidth}
\includegraphics[width = \linewidth, height = 0.12\paperheight]{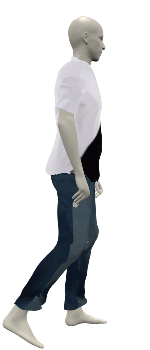}
\end{minipage}
&
\begin{minipage}{0.083\linewidth}
\includegraphics[width = \linewidth, height = 0.12\paperheight]{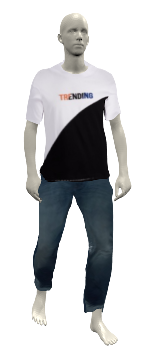}
\end{minipage}
&
\begin{minipage}{0.083\linewidth}
\includegraphics[width = \linewidth, height = 0.12\paperheight]{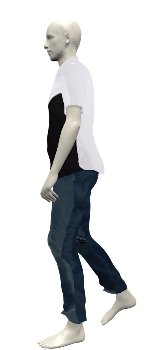}
\end{minipage}
&
\begin{minipage}{0.083\linewidth}
\includegraphics[width = \linewidth, height = 0.12\paperheight]{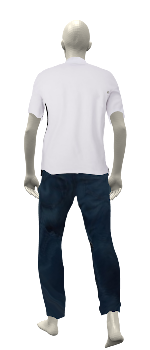}
\end{minipage}
\end{tabular}
\vspace{1mm}
\caption{Textures mapped by Pix2Surf rendered atop SMPL}
\label{fig:results1}
\end{figure*}

\begin{figure*}
\setlength\tabcolsep{0.0pt}
\begin{tabular}{c c c c c c c c c c c c}
\begin{minipage}{0.083\linewidth}
\includegraphics[width = \linewidth, height = 0.06\paperheight ]{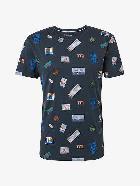}
\\ 
\includegraphics[width = \linewidth, height = 0.06\paperheight]{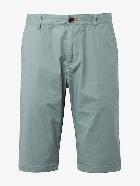}
\end{minipage}
&
\begin{minipage}{0.083\linewidth}
\includegraphics[width = \linewidth, height = 0.06\paperheight ]{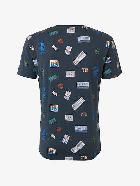}
\\ 
\includegraphics[width = \linewidth, height = 0.06\paperheight]{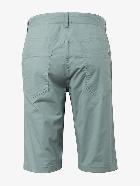}
\end{minipage}
&
\begin{minipage}{0.083\linewidth}
\includegraphics[width = \linewidth, height = 0.12\paperheight]{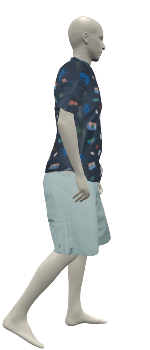}
\end{minipage}
&
\begin{minipage}{0.083\linewidth}
\includegraphics[width = \linewidth, height = 0.12\paperheight]{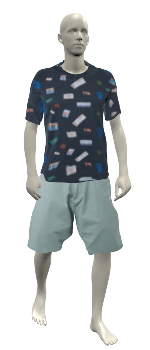}
\end{minipage}
&
\begin{minipage}{0.083\linewidth}
\includegraphics[width = \linewidth, height = 0.12\paperheight]{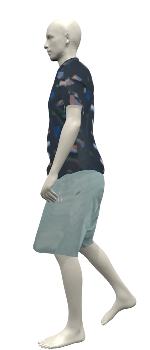}
\end{minipage}
&
\begin{minipage}{0.083\linewidth}
\includegraphics[width = \linewidth, height = 0.12\paperheight]{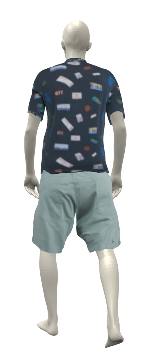}
\end{minipage}
\begin{minipage}{0.083\linewidth}
\includegraphics[width = \linewidth, height = 0.06\paperheight ]{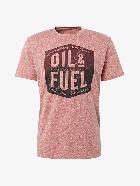}
\\ 
\includegraphics[width = \linewidth, height = 0.06\paperheight]{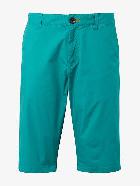}
\end{minipage}
&
\begin{minipage}{0.083\linewidth}
\includegraphics[width = \linewidth, height = 0.06\paperheight ]{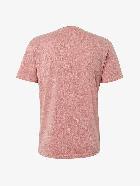}
\\ 
\includegraphics[width = \linewidth, height = 0.06\paperheight]{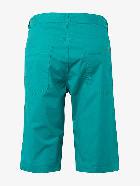}
\end{minipage}
&
\begin{minipage}{0.083\linewidth}
\includegraphics[width = \linewidth, height = 0.12\paperheight]{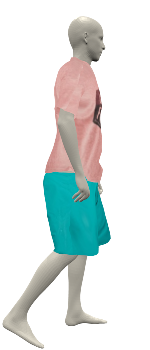}
\end{minipage}
&
\begin{minipage}{0.083\linewidth}
\includegraphics[width = \linewidth, height = 0.12\paperheight]{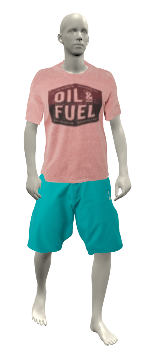}
\end{minipage}
&
\begin{minipage}{0.083\linewidth}
\includegraphics[width = \linewidth, height = 0.12\paperheight]{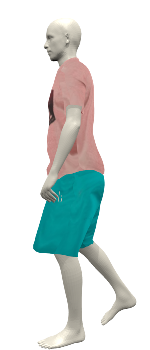}
\end{minipage}
&
\begin{minipage}{0.083\linewidth}
\includegraphics[width = \linewidth, height = 0.12\paperheight]{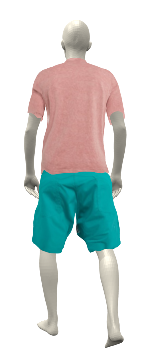}
\end{minipage}
\end{tabular}

\begin{tabular}{c c c c c c c c c c c c}
\begin{minipage}{0.083\linewidth}
\includegraphics[width = \linewidth, height = 0.06\paperheight ]{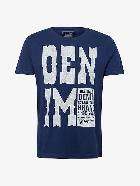}
\\ 
\includegraphics[width = \linewidth, height = 0.06\paperheight]{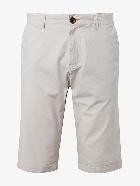}
\end{minipage}
&
\begin{minipage}{0.083\linewidth}
\includegraphics[width = \linewidth, height = 0.06\paperheight ]{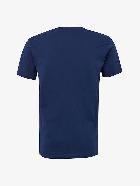}
\\ 
\includegraphics[width = \linewidth, height = 0.06\paperheight]{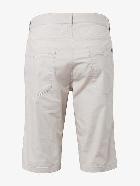}
\end{minipage}
&
\begin{minipage}{0.083\linewidth}
\includegraphics[width = \linewidth, height = 0.12\paperheight]{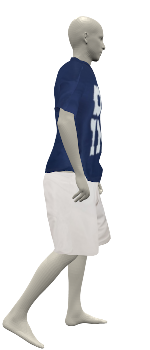}
\end{minipage}
&
\begin{minipage}{0.083\linewidth}
\includegraphics[width = \linewidth, height = 0.12\paperheight]{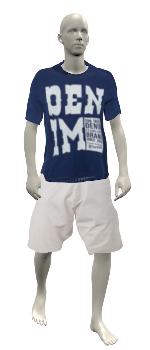}
\end{minipage}
&
\begin{minipage}{0.083\linewidth}
\includegraphics[width = \linewidth, height = 0.12\paperheight]{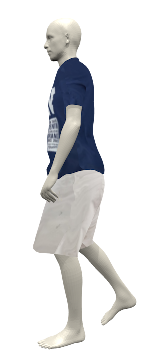}
\end{minipage}
&
\begin{minipage}{0.083\linewidth}
\includegraphics[width = \linewidth, height = 0.12\paperheight]{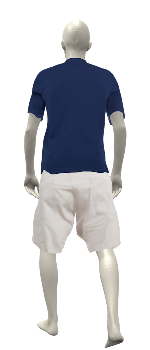}
\end{minipage}
\begin{minipage}{0.083\linewidth}
\includegraphics[width = \linewidth, height = 0.06\paperheight ]{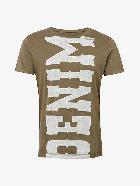}
\\ 
\includegraphics[width = \linewidth, height = 0.06\paperheight]{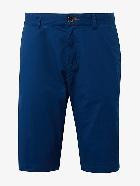}
\end{minipage}
&
\begin{minipage}{0.083\linewidth}
\includegraphics[width = \linewidth, height = 0.06\paperheight ]{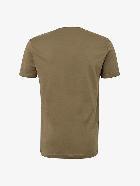}
\\ 
\includegraphics[width = \linewidth, height = 0.06\paperheight]{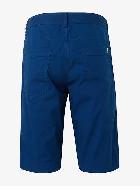}
\end{minipage}
&
\begin{minipage}{0.083\linewidth}
\includegraphics[width = \linewidth, height = 0.12\paperheight]{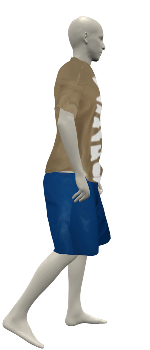}
\end{minipage}
&
\begin{minipage}{0.083\linewidth}
\includegraphics[width = \linewidth, height = 0.12\paperheight]{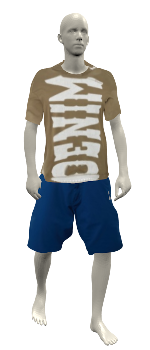}
\end{minipage}
&
\begin{minipage}{0.083\linewidth}
\includegraphics[width = \linewidth, height = 0.12\paperheight]{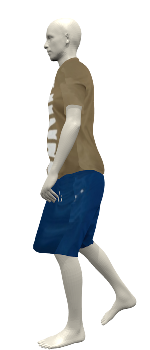}
\end{minipage}
&
\begin{minipage}{0.083\linewidth}
\includegraphics[width = \linewidth, height = 0.12\paperheight]{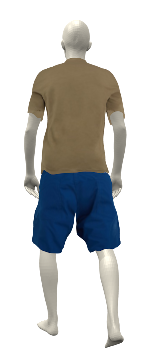}
\end{minipage}
\end{tabular}

\begin{tabular}{c c c c c c c c c c c c}
\begin{minipage}{0.083\linewidth}
\includegraphics[width = \linewidth, height = 0.06\paperheight ]{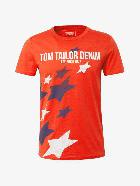}
\\ 
\includegraphics[width = \linewidth, height = 0.06\paperheight]{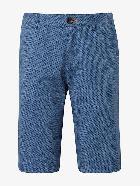}
\end{minipage}
&
\begin{minipage}{0.083\linewidth}
\includegraphics[width = \linewidth, height = 0.06\paperheight ]{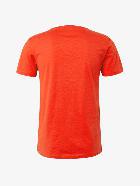}
\\ 
\includegraphics[width = \linewidth, height = 0.06\paperheight]{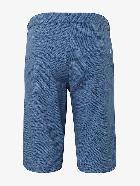}
\end{minipage}
&
\begin{minipage}{0.083\linewidth}
\includegraphics[width = \linewidth, height = 0.12\paperheight]{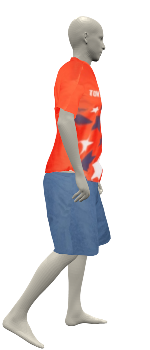}
\end{minipage}
&
\begin{minipage}{0.083\linewidth}
\includegraphics[width = \linewidth, height = 0.12\paperheight]{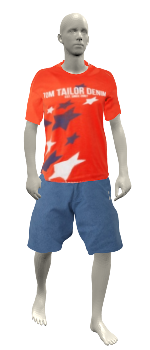}
\end{minipage}
&
\begin{minipage}{0.083\linewidth}
\includegraphics[width = \linewidth, height = 0.12\paperheight]{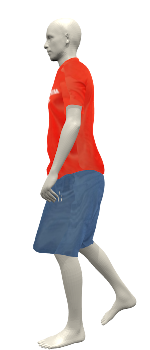}
\end{minipage}
&
\begin{minipage}{0.083\linewidth}
\includegraphics[width = \linewidth, height = 0.12\paperheight]{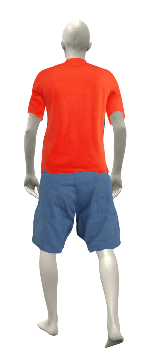}
\end{minipage}
\begin{minipage}{0.083\linewidth}
\includegraphics[width = \linewidth, height = 0.06\paperheight ]{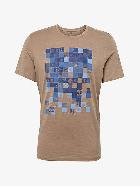}
\\ 
\includegraphics[width = \linewidth, height = 0.06\paperheight]{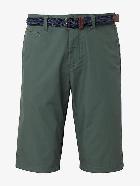}
\end{minipage}
&
\begin{minipage}{0.083\linewidth}
\includegraphics[width = \linewidth, height = 0.06\paperheight ]{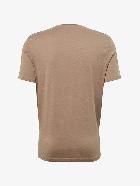}
\\ 
\includegraphics[width = \linewidth, height = 0.06\paperheight]{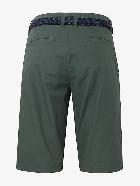}
\end{minipage}
&
\begin{minipage}{0.083\linewidth}
\includegraphics[width = \linewidth, height = 0.12\paperheight]{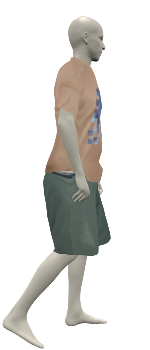}
\end{minipage}
&
\begin{minipage}{0.083\linewidth}
\includegraphics[width = \linewidth, height = 0.12\paperheight]{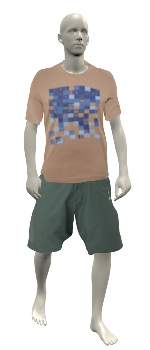}
\end{minipage}
&
\begin{minipage}{0.083\linewidth}
\includegraphics[width = \linewidth, height = 0.12\paperheight]{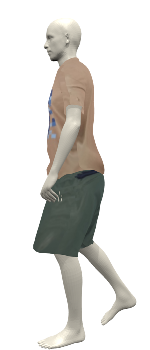}
\end{minipage}
&
\begin{minipage}{0.083\linewidth}
\includegraphics[width = \linewidth, height = 0.12\paperheight]{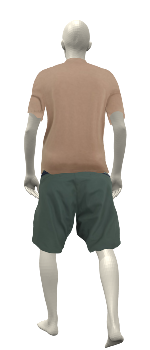}
\end{minipage}
\end{tabular}

\begin{tabular}{c c c c c c c c c c c c}
\begin{minipage}{0.083\linewidth}
\includegraphics[width = \linewidth, height = 0.06\paperheight ]{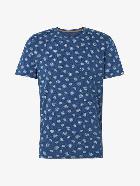}
\\ 
\includegraphics[width = \linewidth, height = 0.06\paperheight]{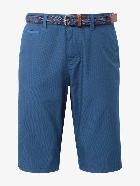}
\end{minipage}
&
\begin{minipage}{0.083\linewidth}
\includegraphics[width = \linewidth, height = 0.06\paperheight ]{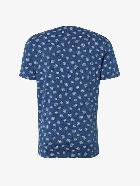}
\\ 
\includegraphics[width = \linewidth, height = 0.06\paperheight]{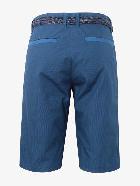}
\end{minipage}
&
\begin{minipage}{0.083\linewidth}
\includegraphics[width = \linewidth, height = 0.12\paperheight]{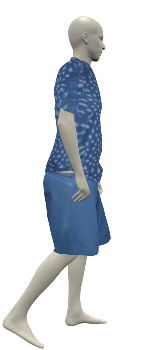}
\end{minipage}
&
\begin{minipage}{0.083\linewidth}
\includegraphics[width = \linewidth, height = 0.12\paperheight]{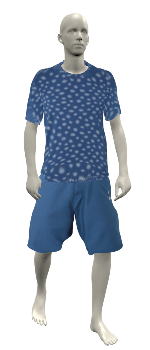}
\end{minipage}
&
\begin{minipage}{0.083\linewidth}
\includegraphics[width = \linewidth, height = 0.12\paperheight]{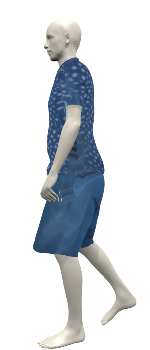}
\end{minipage}
&
\begin{minipage}{0.083\linewidth}
\includegraphics[width = \linewidth, height = 0.12\paperheight]{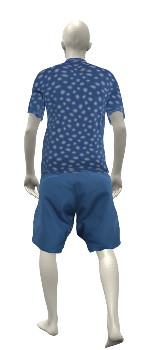}
\end{minipage}
\begin{minipage}{0.083\linewidth}
\includegraphics[width = \linewidth, height = 0.06\paperheight ]{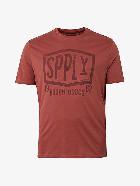}
\\ 
\includegraphics[width = \linewidth, height = 0.06\paperheight]{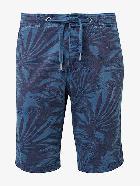}
\end{minipage}
&
\begin{minipage}{0.083\linewidth}
\includegraphics[width = \linewidth, height = 0.06\paperheight ]{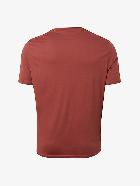}
\\ 
\includegraphics[width = \linewidth, height = 0.06\paperheight]{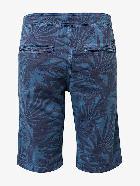}
\end{minipage}
&
\begin{minipage}{0.083\linewidth}
\includegraphics[width = \linewidth, height = 0.12\paperheight]{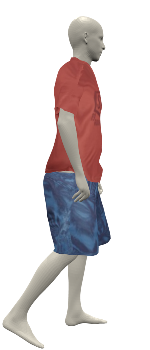}
\end{minipage}
&
\begin{minipage}{0.083\linewidth}
\includegraphics[width = \linewidth, height = 0.12\paperheight]{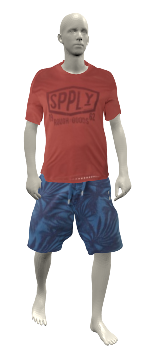}
\end{minipage}
&
\begin{minipage}{0.083\linewidth}
\includegraphics[width = \linewidth, height = 0.12\paperheight]{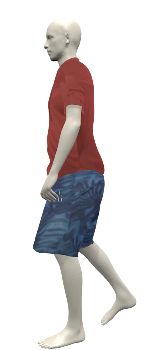}
\end{minipage}
&
\begin{minipage}{0.083\linewidth}
\includegraphics[width = \linewidth, height = 0.12\paperheight]{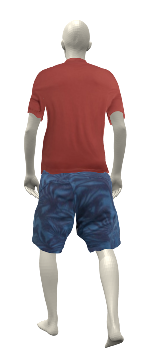}
\end{minipage}
\end{tabular}

\begin{tabular}{c c c c c c c c c c c c}
\begin{minipage}{0.083\linewidth}
\includegraphics[width = \linewidth, height = 0.06\paperheight ]{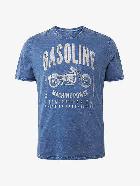}
\\ 
\includegraphics[width = \linewidth, height = 0.06\paperheight]{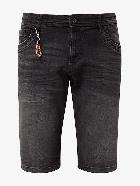}
\end{minipage}
&
\begin{minipage}{0.083\linewidth}
\includegraphics[width = \linewidth, height = 0.06\paperheight ]{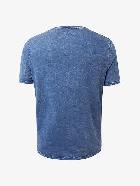}
\\ 
\includegraphics[width = \linewidth, height = 0.06\paperheight]{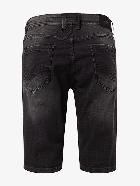}
\end{minipage}
&
\begin{minipage}{0.083\linewidth}
\includegraphics[width = \linewidth, height = 0.12\paperheight]{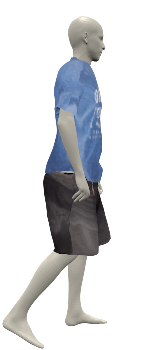}
\end{minipage}
&
\begin{minipage}{0.083\linewidth}
\includegraphics[width = \linewidth, height = 0.12\paperheight]{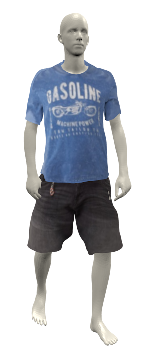}
\end{minipage}
&
\begin{minipage}{0.083\linewidth}
\includegraphics[width = \linewidth, height = 0.12\paperheight]{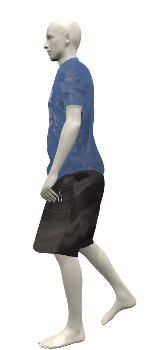}
\end{minipage}
&
\begin{minipage}{0.083\linewidth}
\includegraphics[width = \linewidth, height = 0.12\paperheight]{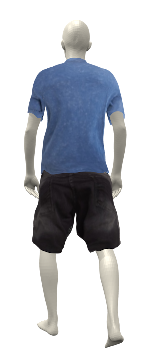}
\end{minipage}
\begin{minipage}{0.083\linewidth}
\includegraphics[width = \linewidth, height = 0.06\paperheight ]{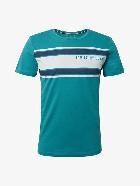}
\\ 
\includegraphics[width = \linewidth, height = 0.06\paperheight]{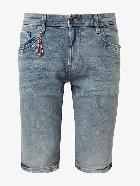}
\end{minipage}
&
\begin{minipage}{0.083\linewidth}
\includegraphics[width = \linewidth, height = 0.06\paperheight ]{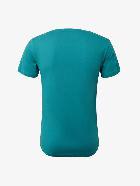}
\\ 
\includegraphics[width = \linewidth, height = 0.06\paperheight]{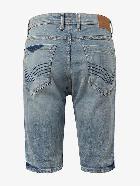}
\end{minipage}
&
\begin{minipage}{0.083\linewidth}
\includegraphics[width = \linewidth, height = 0.12\paperheight]{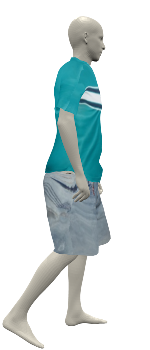}
\end{minipage}
&
\begin{minipage}{0.083\linewidth}
\includegraphics[width = \linewidth, height = 0.12\paperheight]{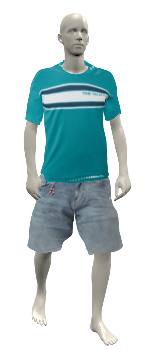}
\end{minipage}
&
\begin{minipage}{0.083\linewidth}
\includegraphics[width = \linewidth, height = 0.12\paperheight]{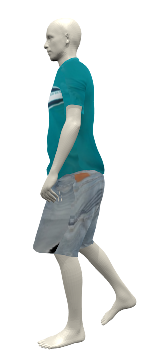}
\end{minipage}
&
\begin{minipage}{0.083\linewidth}
\includegraphics[width = \linewidth, height = 0.12\paperheight]{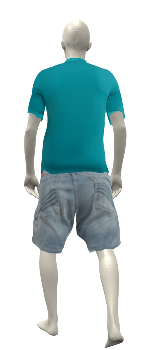}
\end{minipage}
\end{tabular}

\begin{tabular}{c c c c c c c c c c c c}
\begin{minipage}{0.083\linewidth}
\includegraphics[width = \linewidth, height = 0.06\paperheight ]{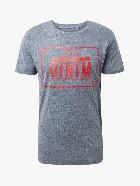}
\\ 
\includegraphics[width = \linewidth, height = 0.06\paperheight]{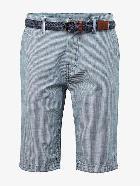}
\end{minipage}
&
\begin{minipage}{0.083\linewidth}
\includegraphics[width = \linewidth, height = 0.06\paperheight ]{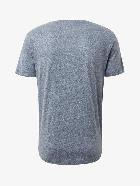}
\\ 
\includegraphics[width = \linewidth, height = 0.06\paperheight]{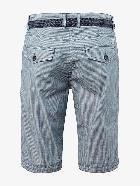}
\end{minipage}
&
\begin{minipage}{0.083\linewidth}
\includegraphics[width = \linewidth, height = 0.12\paperheight]{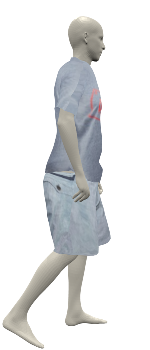}
\end{minipage}
&
\begin{minipage}{0.083\linewidth}
\includegraphics[width = \linewidth, height = 0.12\paperheight]{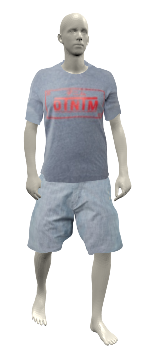}
\end{minipage}
&
\begin{minipage}{0.083\linewidth}
\includegraphics[width = \linewidth, height = 0.12\paperheight]{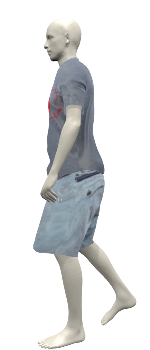}
\end{minipage}
&
\begin{minipage}{0.083\linewidth}
\includegraphics[width = \linewidth, height = 0.12\paperheight]{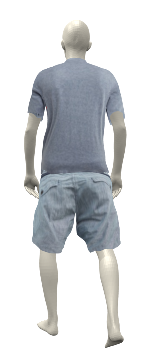}
\end{minipage}
\begin{minipage}{0.083\linewidth}
\includegraphics[width = \linewidth, height = 0.06\paperheight ]{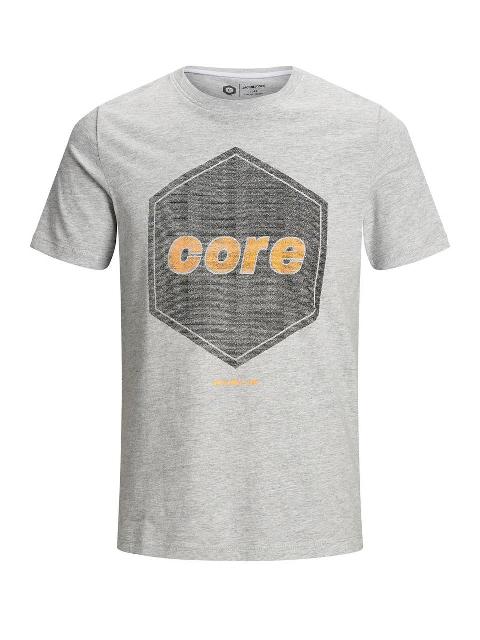}
\\ 
\includegraphics[width = \linewidth, height = 0.06\paperheight]{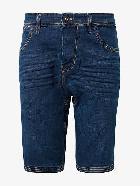}
\end{minipage}
&
\begin{minipage}{0.083\linewidth}
\includegraphics[width = \linewidth, height = 0.06\paperheight ]{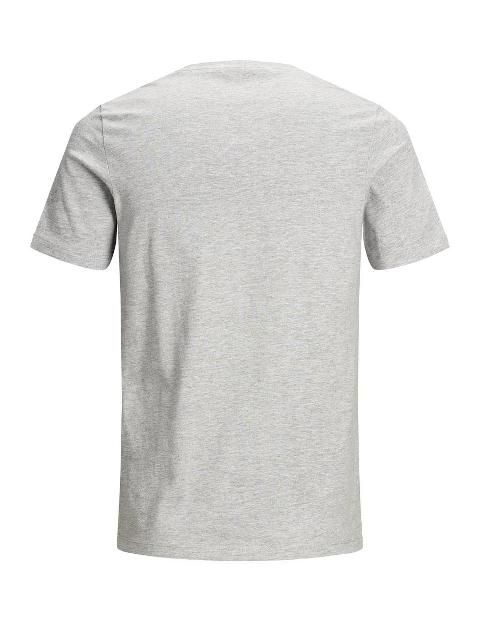}
\\ 
\includegraphics[width = \linewidth, height = 0.06\paperheight]{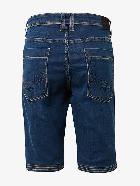}
\end{minipage}
&
\begin{minipage}{0.083\linewidth}
\includegraphics[width = \linewidth, height = 0.12\paperheight]{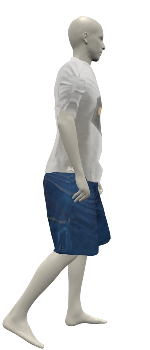}
\end{minipage}
&
\begin{minipage}{0.083\linewidth}
\includegraphics[width = \linewidth, height = 0.12\paperheight]{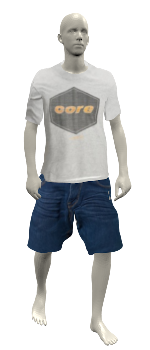}
\end{minipage}
&
\begin{minipage}{0.083\linewidth}
\includegraphics[width = \linewidth, height = 0.12\paperheight]{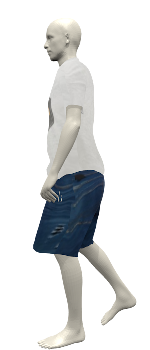}
\end{minipage}
&
\begin{minipage}{0.083\linewidth}
\includegraphics[width = \linewidth, height = 0.12\paperheight]{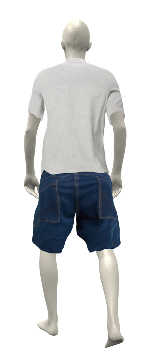}
\end{minipage}
\end{tabular}
\vspace{1mm}
\caption{Textures mapped by Pix2Surf rendered atop SMPL}
\label{fig:results2}
\end{figure*}

In Fig. \ref{fig:results1} and Fig. \ref{fig:results2} more textures mapped by Pix2Surf are displayed atop SMPL. 

\begin{figure*}
    \centering
    \includegraphics[width = \linewidth]{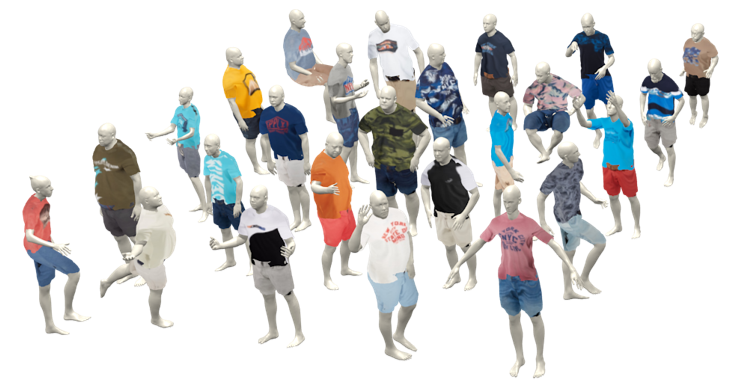}
    \caption{Once textures are mapped to their corresponding texture maps, they can be rendered atop SMPL in different shapes and poses}
    \label{fig:results_shape}
\end{figure*}

Once the garment textures are mapped to their corresponding texture maps, they can be rendered atop SMPL in different shapes and poses as seen in Fig. \ref{fig:results_shape}. 

\end{document}